\documentclass[lettersize,journal]{IEEEtran}
\usepackage{amsmath,amsfonts}
\usepackage{algorithmic}
\usepackage{algorithm}
\usepackage{array}
\usepackage[caption=false,font=normalsize,labelfont=sf,textfont=sf]{subfig}
\usepackage{textcomp}
\usepackage{stfloats}
\usepackage{url}
\usepackage{verbatim}
\usepackage{graphicx}
\usepackage{cite}

\usepackage{hyperref}
\usepackage{xspace}
\usepackage[capitalise]{cleveref}
\usepackage{booktabs}
\usepackage{xcolor}
\usepackage{amsmath,bm}
\usepackage{multirow}

\newcommand{\mainEqAvgPoss}{Eq.~(4) }

\newcommand{\extension}[1]{{#1}}

\newcommand{\extensionvtwo}[1]{{#1}}
\newcommand{\todo}[1]{}
\newcommand{\sma}[1]{{#1}}

\makeatletter
\DeclareRobustCommand\onedot{\futurelet\@let@token\@onedot}
\def\@onedot{\ifx\@let@token.\else.\null\fi\xspace}
\def\eg{\emph{e.g}\onedot} 
\def\ie{\emph{i.e}\onedot} 
 
\def\etc{\emph{etc}\onedot} 
 
\def\etal{\emph{et al}\onedot}
\makeatother

\newsavebox\CBox
\def\textBF#1{\sbox\CBox{#1}\resizebox{\wd\CBox}{\ht\CBox}{\textbf{#1}}}
\newcommand{\bestscore}[1]{\textcolor{black}{\textBF{#1}}}

\newcommand{\fakescore}[1]{\textcolor{gray}{#1}}

\begin{document}

\title{\extension{Velocity Disambiguation for Video Frame Interpolation}}

\author{
Zhihang Zhong,
\thanks{$^{\dagger}$ denotes corresponding authors.}
\thanks{Z.~Zhong is with the School of Artificial Intelligence, Shanghai Jiao Tong University. Y.~Zhang is with Cornell University. W.~Wang, X.~Sun and Y.~Qiao. are with Shanghai Artificial Intelligence Laboratory. G.~Krishnan is with OtoNexus
Medical Technologies. S.~Ma and J.~Wang. are with Snap Inc.}
\thanks{Part of the work was done while
Z. Zhong and G.~Krishnan  were at Snap.}
Yiming Zhang,
Wei Wang,
Xiao Sun,
Yu Qiao,
Gurunandan Krishnan,
\\Sizhuo Ma$^{\dagger}$,
and Jian Wang$^{\dagger}$
}

\markboth{Journal of \LaTeX\ Class Files,~Vol.~14, No.~8, August~2021}%
{Shell \MakeLowercase{\textit{et al.}}: A Sample Article Using IEEEtran.cls for IEEE Journals}

\maketitle

\begin{abstract}
Existing video frame interpolation (VFI) methods blindly predict where each object is at a specific timestep $t$ (``time indexing''), which struggles to predict precise object movements. Given two images of a baseball, there are infinitely many possible trajectories: accelerating or decelerating, straight or curved. This often results in blurry frames as the method averages out these possibilities. Instead of forcing the network to learn this complicated time-to-location mapping implicitly together with predicting the frames, we provide the network with an explicit hint on how far the object has traveled between start and end frames, a novel approach termed ``distance indexing''. This method offers a clearer learning goal for models, reducing the uncertainty tied to object speeds. We further observed that, even with this extra guidance, objects can still be blurry especially when they are equally far from both input frames (\ie, halfway in-between), due to the directional ambiguity in long-range motion. 
To solve this, we propose an iterative reference-based estimation strategy that breaks down a long-range prediction into several short-range steps. When integrating our plug-and-play strategies into state-of-the-art learning-based models, they exhibit markedly sharper outputs and superior perceptual quality in arbitrary time interpolations, using a uniform distance indexing map in the same format as time indexing \sma{without requiring extra computation}.
\sma{Furthermore, we demonstrate that if additional latency is acceptable, a continuous map estimator can be employed to compute a pixel-wise dense distance indexing using multiple nearby frames. Combined with efficient multi-frame refinement, this extension can further disambiguate complex motion, thus enhancing performance both qualitatively and quantitatively. Additionally, the ability to manually specify distance indexing allows for independent temporal manipulation of each object, providing a novel tool for video editing tasks such as re-timing.}
The code is available at \href{https://zzh-tech.github.io/InterpAny-Clearer/}{https://zzh-tech.github.io/InterpAny-Clearer/}.
\end{abstract}

\begin{IEEEkeywords}
Video frame interpolation, Temporal super-resolution, Disambiguation, Video editing
\end{IEEEkeywords}

\section{Introduction}
\label{sec:introduction}
\IEEEPARstart{V}{ideo} frame interpolation (VFI) plays a crucial role in creating slow-motion videos~\cite{bao2019depth}, video generation~\cite{ho2022imagen}, prediction ~\cite{wu2022optimizing}, and compression~\cite{wu2018video}. Directly warping the starting and ending frames using the optical flow between them can only model linear motion, which often diverges from actual motion paths, leading to artifacts such as holes. To solve this, learning-based methods have emerged as leading solutions to VFI, which aim to develop a model, represented as $\mathcal{F}$, that uses a starting frame $I_0$ and an ending frame $I_1$ to generate a frame for a given timestep, described by:
\begin{equation}
\label{eq:deterministic}
I_{t} = \mathcal{F}\left(I_0, I_1, t\right).
\end{equation}
Two paradigms have been proposed: In fixed-time interpolation~\cite{liu2017video,bao2019depth}, the model only takes the two frames as input and always tries to predict the frame at $t=0.5$. In arbitrary-time interpolation~\cite{jiang2018super,huang2022real}, the model is further given a user-specified timestep $t\in[0,1]$, which is more flexible at predicting multiple frames in-between.

Yet, in both cases, the unsampled blank between the two frames, such as the motion between a ball's starting and ending points, presents infinite possibilities. The velocities of individual objects within these frames remain undefined, introducing a \emph{velocity ambiguity}, a myriad of plausible time-to-location mappings during training. 
\sma{Incorporating additional neighboring frames as input \cite{xu2019quadratic} can partially restrict the solution space by imposing constraints on motion trajectories, but it cannot fully resolve the ambiguity.}
We observed that velocity ambiguity is a primary obstacle hindering the advancement of learning-based VFI: Models trained using aforementioned \emph{time indexing} receive identical inputs with differing supervision signals during training. As a result, they tend to produce blurred and imprecise interpolations, as they average out the potential outcomes.

Could an alternative indexing method minimize such conflicts?
One straightforward option is to provide the optical flow at the target timestep as an explicit hint on object motion. 
However, this information is unknown at inference time, which has to be approximated by the optical flow between $I_0$ and $I_1$, scaled by the timestep. This requires running optical flow estimation on top of VFI, which may increase the computational complexity and enforce the VFI algorithm to rely on the explicitly computed but approximate flow.
Instead, we propose a more flexible \emph{distance indexing} approach. In lieu of an optical flow map, we employ a \emph{distance ratio} map $D_t$, where each pixel denotes \emph{how far the object has traveled between start and end frames}, within a normalized range of $[0,1]$,
\begin{equation}
\boxed{
\label{eq:ours}
I_{t} = \mathcal{F}\left(I_0, I_1, \text{motion hint}\right) \; \Rightarrow \; I_{t} = \mathcal{F}\left(I_0, I_1, D_t\right).
}
\end{equation}
{\bf During training, $D_t$ is derived from optical flow ratios computed from ground-truth frames.}
{\bf During inference, it is sufficient to provide a uniform map as input}, in the exactly same way as time indexing methods, \ie, $D_t(x,y)=t,\,\forall x,y$.
However, the semantics of this indexing map have shifted from an uncertain timestep map to a more deterministic motion hint.
Through distance indexing, we effectively solve the one-to-many time-to-position mapping dilemma, fostering enhanced convergence and interpolation quality.

\begin{figure*}[htbp]
    \begin{center}
        \includegraphics[width=\linewidth]{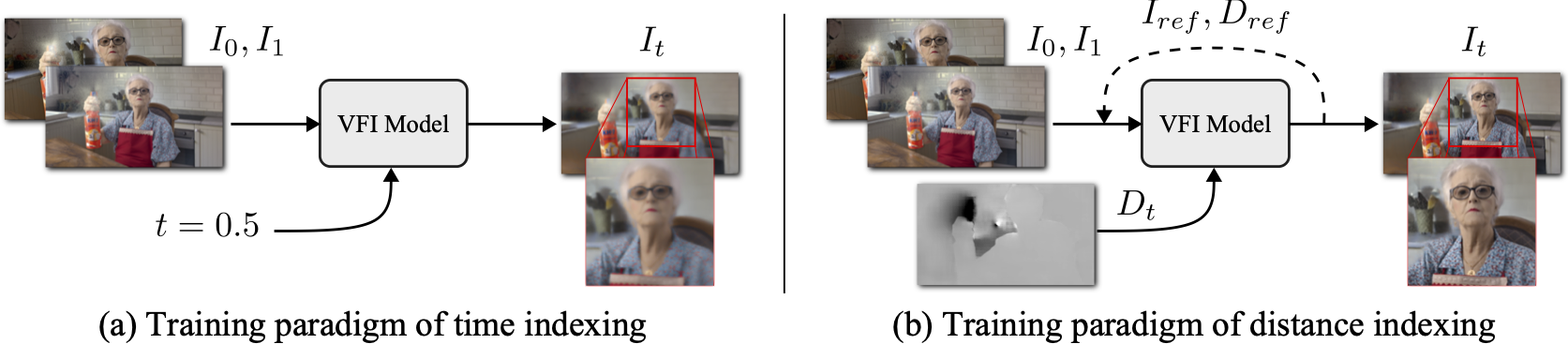}
    \end{center}
\caption{\extension{Comparison of time indexing and distance indexing training paradigms. (a) Time indexing uses the starting frame $I_0$, ending frame $I_1$, and a scalar variable $t$ as inputs. (b) Distance indexing replaces the scalar with a distance map $D_t$ and optionally incorporates iterative reference-based estimation $\left(I_{ref}, D_{ref}\right)$ to address velocity ambiguity, resulting in a notably sharper prediction.}}
\label{fig:teaser}
\end{figure*}

\extension{Although distance indexing addresses the scalar \emph{speed ambiguity}, the \emph{directional ambiguity} of motion remains a challenge. Empirically we found that, although learning-based video frame interpolation can handle minor directional uncertainty at a small timestep, the ambiguity becomes pronounced at the temporal midway between the two input frames, \ie, t=0.5, as illustrated in \cref{fig:disambiguate}(b). Inspired by iterative inference paradigms in optical flow~\cite{teed2020raft} and image generation~\cite{rombach2022high}, we introduce an iterative reference-based estimation strategy. 
Rather than estimating the full motion field at once, our approach decomposes the problem into incremental distance steps. By propagating estimates from nearby to farther points, we constrain the search space at each iteration, thereby minimizing directional uncertainty and enhancing synthesis quality.}

Our approach addresses challenges that are not bound to specific network architectures. Indeed, it can be applied as a plug-and-play strategy that requires only modifying the input channels for each model, as demonstrated in \cref{fig:teaser}. We conducted extensive experiments on four existing VFI methods to validate the effectiveness of our approach, which produces frames of markedly improved perceptual quality. Moreover, instead of using a uniform map, it is also possible to use a spatially-varying 2D map as input to manipulate the motion of objects. Paired with state-of-the-art segmentation models such as Segment Anything Model (SAM)~\cite{kirillov2023segment}, this empowers users to freely control the interpolation of any object, \eg, making certain objects backtrack in time. 

\sma{When using more than two input frames \cite{xu2019quadratic}, nearby frames offer additional constraints that facilitate the computation of a pixel-wise dense distance map. The methods discussed so far employ a uniform distance map because a deterministic distance map cannot be derived from only two frames. Inspired by continuous parametric optical flow estimation \cite{luo2024continuous}, we utilize cubic B-splines and neural ordinary differential equations to estimate a dense distance map from four input frames, which improves our model's performance across both perceptual and pixel-centric metrics.
Furthermore, we make a trainable copy of the original interpolator architecture to refine the initial two-frame interpolation results using information from two additional frames $I_{-1}$ and $I_2$. This multi-frame refiner module is intended to fully harness the potential of additional frames, thereby enhancing multi-frame interpolation quality.
}

In summary, our key contributions are: 
1) Proposing distance indexing and iterative reference-based estimation to address the velocity ambiguity and enhance the capabilities of arbitrary time interpolation models; 
2) Presenting an unprecedented manipulation method that allows for customized interpolation of any object. 
\extension{3) Adopting a continuous distance map estimator and proposing a multi-frame fusion architecture to enhance interpolation quality across both perceptual and pixel-centric metrics.}

\sma{A preliminary version of this work was presented in \cite{zhong2025clearer}, where we focus on using a uniform distance map during inference because an accurate, pixel-wise distance map cannot be reliably calculated from two frames. Although using a uniform distance map produced plausible results with better perceptual quality, the predictions did not align with the ground truth, resulting in lower performance on metrics such as PSNR and SSIM. In this paper, we address this limitation by using multiple frames (more than two) as input and introduce a continuous distance map estimator that approximates the map from nearby frames. We also present a simple yet effective multi-frame refiner architecture for video frame interpolation. Extensive experiments demonstrate that this approach significantly enhances performance.}

\section{Related Work}
\subsection{Video frame interpolation}
\subsubsection{General overview} 
Numerous VFI solutions rely on optical flows to predict latent frames. Typically, these methods warp input frames forward or backward using flow calculated by off-the-shelf networks like ~\cite{sun2018pwc,dosovitskiy2015flownet,ilg2017flownet,teed2020raft} or self-contained flow estimators like~\cite{huang2022real, zhang2023extracting, li2023amt}. Networks then refine the warped frame to improve visual quality. SuperSlomo~\cite{jiang2018super} uses a linear combination of bi-directional flows for intermediate flow estimation and backward warping. DAIN~\cite{bao2019depth} introduces a depth-aware flow projection layer for advanced intermediate flow estimation. AdaCoF~\cite{lee2020adacof} estimates kernel weights and offset vectors for each target pixel, while BMBC~\cite{park2020bmbc} and ABME~\cite{park2021asymmetric} refine optical flow estimation. Large motion interpolation is addressed by XVFI~\cite{sim2021xvfi} through a recursive multi-scale structure. 
\extension{VFIFormer~\cite{lu2022video} employs Transformer architectures to model long-range pixel correlations, while VFIMamba~\cite{zhang2024vfimamba} adopts a Mamba-based architecture for efficient sequence modeling.} IFRNet~\cite{kong2022ifrnet}, RIFE~\cite{huang2022real}, and UPR-Net~\cite{jin2023unified} employ efficient pyramid network designs for high-quality, real-time interpolation, with IFRNet and RIFE using leakage distillation losses for flow estimation. 
Recently, more advanced network modules and operations are proposed to push the upper limit of VFI performance, such as the transformer-based bilateral motion estimator of BiFormer~\cite{park2023biformer}, a unifying operation of EMA-VFI~\cite{zhang2023extracting} to explicitly disentangle motion and appearance information, and bi-directional correlation volumes for all pairs of pixels of AMT~\cite{li2023amt}. 
On the other hand, SoftSplat~\cite{niklaus2020softmax} and M2M~\cite{hu2022many} actively explore the forward warping operation for VFI. 
\extension{To further improve perceptual clarity in video frame interpolation, FILM~\cite{reda2022film} introduces perceptual losses during training, while uncertainty-aware and adaptive interpolation methods have also been proposed to better handle ambiguous regions~\cite{plack2023frame,briedis2023kernel}.}
\extension{More recently, diffusion-based video frame interpolators, such as LDMVFI \cite{danier2024ldmvfi} and SVDKFI \cite{wang2024generative}, have leveraged generative priors to enhance visual quality but still face challenges such as  high computational costs and slow runtimes.}

Other contributions to VFI come from various perspectives. For instance, Xu~\etal~\cite{xu2019quadratic,hu2024iq} leverage acceleration information from nearby frames, VideoINR \cite{chen2022videoinr} is the first to employ an implicit neural representation, and Lee~\etal~\cite{lee2023exploring} explore and address discontinuity in video frame interpolation using figure-text mixing data augmentation and a discontinuity map. Flow-free approaches have also attracted interest. SepConv~\cite{niklaus2017video} integrates motion estimation and pixel synthesis, CAIN~\cite{choi2020channel} employs the PixelShuffle operation with channel attention, and FLAVR~\cite{kalluri2023flavr} utilizes 3D space-time convolutions. Additionally, specialized interpolation methods for anime, which often exhibit minimal textures and exaggerated motion, are proposed by AnimeInterp~\cite{siyao2021deep} and Chen~\etal~\cite{chen2022improving}.
On the other hand, motion induced blur~\cite{shen2020blurry,zhong2022animation,zhong2023blur}, shutter mode~\cite{fan2021inverting,zhong2022bringing,ji2023rethinking}, and event camera~\cite{tulyakov2021time,lin2023event} are also exploited to achieve VFI. \extension{For a more comprehensive overview of recent advances in video frame interpolation, we refer readers to the survey by Kye~\etal~\cite{kye2025acevfi}.}

\begin{figure*}[htbp]
    \begin{center}
    \includegraphics[width=.9\linewidth]{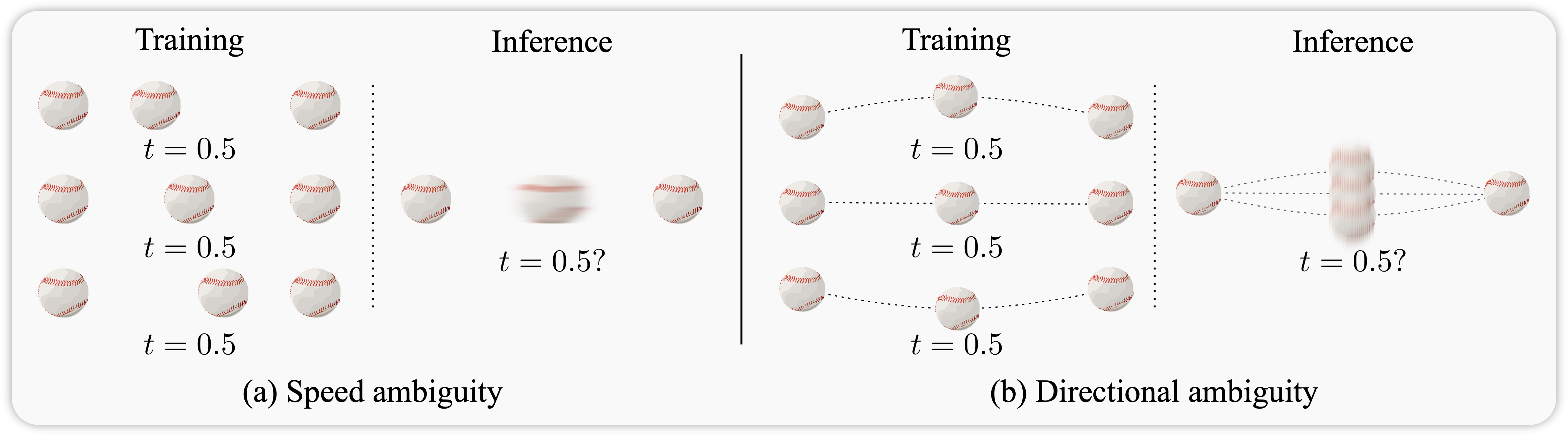}
    \end{center}
\caption{Velocity ambiguity. (a) Speed ambiguity. (b) Directional ambiguity.}
\label{fig:ambiguity}
\end{figure*}

\begin{figure*}[htbp]
    \begin{center}
        \includegraphics[width=.9\linewidth]{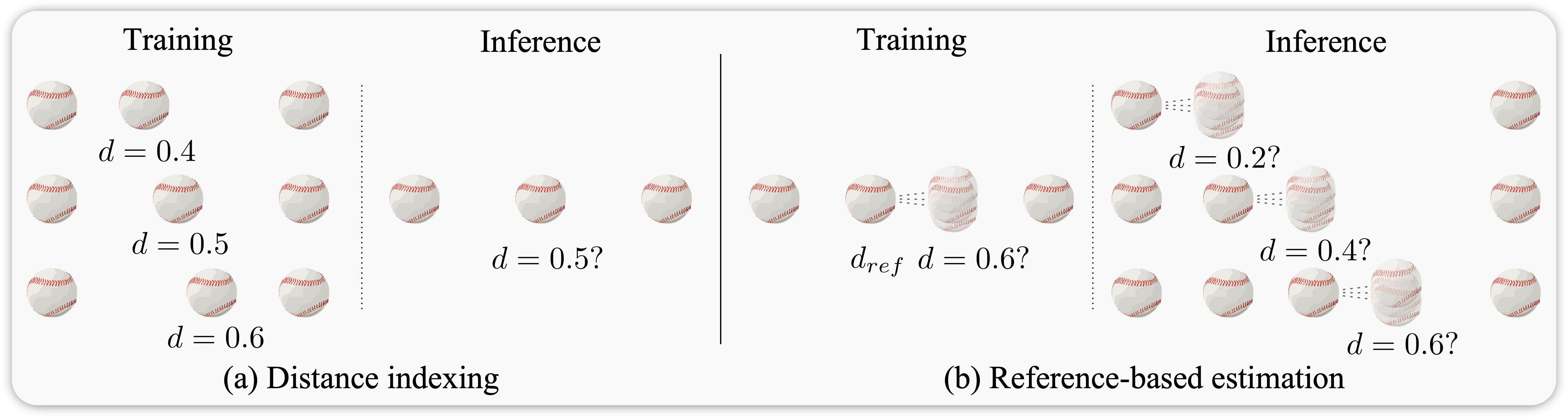}
    \end{center}
\caption{Disambiguation strategies for velocity ambiguity. (a) Distance indexing. (b) Iterative reference-based estimation.}
\label{fig:disambiguate}
\end{figure*}

\subsubsection{Learning paradigms}
One major thread of VFI methods train networks on triplet of frames, always predicting the central frame. Iterative estimation is used for interpolation ratios higher than $\times$2. This \emph{fixed-time} method often accumulates errors and struggles with interpolating at arbitrary continuous timesteps. Hence, models like SuperSloMo~\cite{jiang2018super}, DAIN~\cite{bao2019depth}, BMBC~\cite{park2020bmbc}, EDSC~\cite{cheng2021multiple}, RIFE~\cite{huang2022real}, IFRNet~\cite{kong2022ifrnet}, EMA-VFI~\cite{zhang2023extracting}, and AMT~\cite{li2023amt} have adopted an \emph{arbitrary time} interpolation paradigm. While theoretically superior, the arbitrary approach faces challenges of more complicated time-to-position mappings due to the velocity ambiguity, resulting in blurred results. This study addresses velocity ambiguity in arbitrary time interpolation and offers solutions.

Prior work by Zhou~\etal~\cite{zhou2023exploring} identified motion ambiguity and proposed a texture consistency loss to implicitly ensure interpolated content resemblance to given frames. In contrast, we explicitly address velocity ambiguity and propose solutions. These innovations not only enhance the performance of arbitrary time VFI models but also offer advanced manipulation capabilities. 
\sma{Additionally, we demonstrate that leveraging information from nearby frames through a multi-frame refiner module, combined with continuous indexing map estimation, can further improve interpolation quality.}

\subsubsection{Segment anything}
The emergence of Segment Anything Model (SAM)~\cite{kirillov2023segment} has marked a significant advancement in the realm of zero-shot segmentation, enabling numerous downstream applications including video tracking and segmentation~\cite{yang2023track}, breakthrough mask-free inpainting techniques~\cite{yu2023inpaint}, and interactive image description generation~\cite{wang2023caption}. 
By specifying the distance indexing individually for each segment, this work introduces a pioneering application to this growing collection: Manipulated Interpolation of Anything.

\begin{figure*}[htbp]
    \begin{center}
    \includegraphics[width=\linewidth]{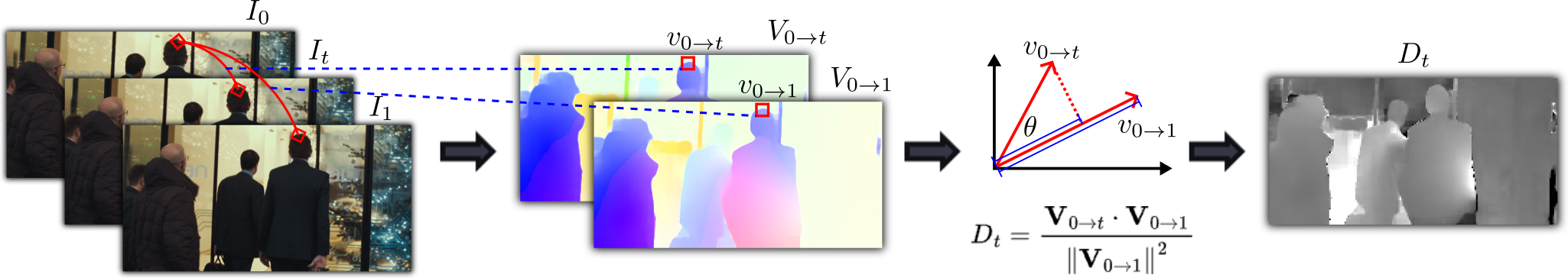}
    \end{center}
\caption{\extension{Calculation of distance map for distance indexing. $V_{0\to t}$ is the estimated optical flow from $I_0$ to $I_t$ by RAFT~\cite{teed2020raft}, and $V_{0\to 1}$ is the optical flow from $I_0$ to $I_1$.}}
\label{fig:d_indexing}
\end{figure*}

\section{Velocity Ambiguity}
\label{sec:ambiguity}

In this section, we begin by revisiting the time indexing paradigm. We then outline the associated velocity ambiguity, which encompasses both speed and directional ambiguities.

\cref{fig:ambiguity} (a) shows the example of a horizontally moving baseball. Given a starting frame and an ending frame, along with a time indexing variable $t=0.5$, the goal of a VFI model is to predict a latent frame at this particular timestep, in accordance with \cref{eq:deterministic}.

Although the starting and ending positions of the baseball are given, its location at $t=0.5$ remains ambiguous due to an unknown speed distribution: The ball can be accelerating or decelerating, resulting in different locations. This ambiguity introduces a challenge in model training as it leads to multiple valid supervision targets for the identical input. Contrary to the deterministic scenario illustrated in \cref{eq:deterministic}, \extension{the VFI function $\mathcal{F}$ is actually tasked with generating a sampled sequence of plausible frames within the \emph{distribution} derived from the same input frames and time indexing.} This can be expressed as:
\begin{equation}
\label{eq:uncertainty}
    \left\{I_t^1, I_t^2, \ldots, I_t^n\right\} = \mathcal{F}(I_0, I_1, t),
\end{equation}
where $n$ is the number of plausible frames. Empirically, the model, when trained with this ambiguity, tends to produce a weighted average of possible frames during inference. While this minimizes the loss during training, it results in blurry frames that are perceptually unsatisfying to humans, as shown in \cref{fig:teaser} (a). This blurry prediction $\hat{I}_t$ can be considered as an average over all the possibilities if an $L_2$ loss is used:
\begin{equation}
\label{eq:avg_poss}
    \hat{I}_t = \mathbb{E}_{I_t \sim \mathcal{F}(I_0, I_1, t)}[I_t].\qquad\text{(See details in Appendix)}
\end{equation}
For other losses, \cref{eq:avg_poss} no longer holds, but we empirically observe that the model still learns an aggregated mixture of training frames which results in blur (RIFE~\cite{huang2022real} and EMA-VFI~\cite{zhang2023extracting}: Laplacian loss, \ie, L1 loss between the Laplacian pyramids of image pairs; IFRNet~\cite{kong2022ifrnet} and AMT~\cite{li2023amt}: Charbonnier loss).

Indeed, not only the speed but also the direction of motion remains indeterminate, leading to what we term as ``directional ambiguity.'' This phenomenon is graphically depicted in \cref{fig:ambiguity} (b). This adds an additional layer of complexity in model training and inference. We collectively refer to speed ambiguity and directional ambiguity as velocity ambiguity.

So far, we have been discussing the ambiguity for the fixed time interpolation paradigm, in which $t$ is set by default to $0.5$. For arbitrary time interpolation, the ambiguity becomes more pronounced: Instead of predicting a single timestep, the network is expected to predict a continuum of timesteps between 0 and 1, each having a multitude of possibilities. This further complicates learning. Moreover, this ambiguity is sometimes referred to as \emph{mode averaging}, which has been studied in other domains~\cite{wang2018style}. See Appendix for details. 

\section{Disambiguation Strategies}
\label{sec:disambiguate}

In this section, we introduce two innovative strategies, namely distance indexing and iterative reference-based estimation, aimed at addressing the challenges posed by the velocity ambiguity. Designed to be plug-and-play, these strategies can be seamlessly integrated into any existing VFI models without necessitating architectural modifications, as shown in \cref{fig:teaser} (b).

In traditional time indexing, models intrinsically deduce an uncertain time-to-location mapping, represented as $\mathcal{D}$:
\begin{equation}
    I_t = \mathcal{F}(I_0, I_1, \mathcal{D}(t)).
\end{equation}
This brings forth the question: Can we guide the network to interpolate more precisely without relying on the ambiguous mapping $\mathcal{D}(t)$ to decipher it independently? 
To address this, we introduce a strategy to diminish speed uncertainty by directly specifying a distance ratio map ($D_t$) instead of the uniform timestep map. This is termed as distance indexing. Consequently, the model sidesteps the intricate process of deducing the time-to-location mapping:
\begin{equation}
    I_t = \mathcal{F}(I_0, I_1, D_t).
\end{equation}

\subsection{Distance indexing}
We utilize an off-the-shelf optical flow estimator, RAFT~\cite{teed2020raft}, to determine the pixel-wise distance map, as shown in \cref{fig:d_indexing}. Given an image triplet $\left\{I_0, I_1, I_t\right\}$, we first calculate the optical flow from $I_0$ to $I_t$, denoted as $\mathbf{V}_{0\to t}$, and from $I_0$ to $I_1$ as $\mathbf{V}_{0\to 1}$. 
At each pixel $(x, y)$, we project the motion vector $\mathbf{V}_{0\to t}(x,y)$ onto $\mathbf{V}_{0\to 1}(x,y)$. The distance map is then defined as the ratio between the projected $\mathbf{V}_{0\to t}(x,y)$ and $\mathbf{V}_{0\to 1}(x,y)$:
\extension{
\begin{equation}
    D_t(x,y) = \frac{\mathbf{V}_{0\to t}(x,y) \cdot \mathbf{V}_{0\to 1}(x,y)}{\left\Vert \mathbf{V}_{0\to 1}(x,y) \right\Vert^2},
\label{eq:dt}
\end{equation}
}where $\theta$ denotes the angle between the two. 
By directly integrating $D_t$, the network achieves a clear comprehension of distance during its training phase, subsequently equipping it to yield sharper frames during inference, as showcased in \cref{fig:disambiguate} (a).

During inference, the algorithm does not have access to the exact distance map since $I_t$ is unknown. In practice, we notice it is usually sufficient to provide a uniform map $D_t=t$, similar to time indexing. Physically this encourages the model to move each object at constant speeds along their trajectories. We observe that constant speed between frames is a valid approximation for many real-world situations. In \cref{sec:experiment}, we show that even though this results in pixel-level misalignment with the ground-truth frames, it achieves significantly higher perceptual scores and is strongly preferred in the user study. Precise distance maps can be computed from multiple frames, which can potentially further boost the performance. See a detailed discussion in Appendix.

\subsection{Iterative reference-based estimation}
While distance indexing addresses speed ambiguity, it omits directional information, leaving directional ambiguity unresolved. Our observations indicate that, even with distance indexing, frames predicted at greater distances from the starting and ending frames remain not clear enough due to this ambiguity. To address this, we propose an iterative reference-based estimation strategy, which divides the complex interpolation for long distances into shorter, easier steps. This strategy enhances the traditional VFI function, $\mathcal{F}$, by incorporating a reference image, $I_{\text{ref}}$, and its corresponding distance map, $D_{\text{ref}}$. Specifically, the network now takes the following channels as input:
\begin{equation}
    I_t = \mathcal{F}(I_0, I_1, D_t, I_{\text{ref}}, D_{\text{ref}}).
\end{equation}

In the general case of $N$ steps, the process of iteration is as follows:
\begin{equation}
    I_{(i+1)t/N} = \mathcal{F}(I_0, I_1, D_{(i+1)t/N}, I_{t/N}, D_{t/N}),
\end{equation}
where $i\in\{0,1,\dots,N-1\}$. For example, if we break the estimation of a remote step $t$ into two steps:
\begin{align}
    I_{t/2} = \mathcal{F}(I_0, I_1, D_{t/2}, I_{0}, D_{0}),\\
    I_{t} = \mathcal{F}(I_0, I_1, D_{t}, I_{t/2}, D_{t/2}).
\end{align}
Importantly, in every iteration, we consistently use the starting and ending frames as reliable appearance references, preventing divergence of uncertainty in each step. By dividing a long step into shorter steps, the uncertainty in each step is reduced, as shown in \cref{fig:disambiguate} (b). While fixed time models also employ an iterative method in a bisectioning way during inference, our strategy progresses from near to far, ensuring more deterministic trajectory interpolation. This reduces errors and uncertainties tied to a single, long-range prediction. \textbf{See more on the rationale for solving ambiguities in Appendix.}

\begin{figure}[!t]
    \begin{center}
    \includegraphics[width=\linewidth]{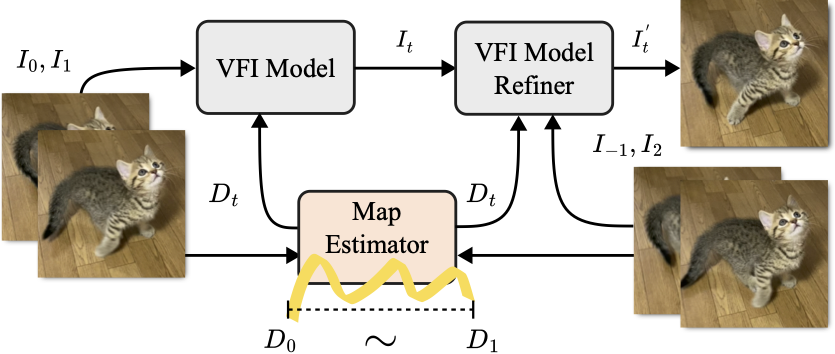}
    \end{center}
\caption{\extension{Multi-frame fusion architecture with continuous map estimator.}}
\label{fig:multi-frame}
\end{figure}

\section{\sma{Leveraging Information from Nearby Frames}}
\label{sec:nearby}
\extension{In this section, we first introduce a continuous indexing map estimator to achieve pixel-wise distance map estimation. Next, we demonstrate how to reuse the original interpolation architecture to fuse nearby frames for enhancing interpolation quality. The complete framework is illustrated in~\cref{fig:multi-frame}.} 

\subsection{\sma{Pixel-wise distance map estimation}}
\extension{To achieve a continuous indexing map estimator, we adopt the pretrained model introduced in CPFlow~\cite{luo2024continuous}. 
CPFlow takes a sequence of images as input \sma{(four in our case)} and predicts the optical flow from the initial frame to any arbitrary timestamp within the sequence. Specifically, it models each pixel's motion trajectory using cubic B-splines, enabling dense and temporally continuous flow estimation. Given normalized time \( t \in [0, 1] \), the displacement of a pixel is defined as:}
\begin{equation}
    \sma{
    \mathbf{V}_{0\to t} = \sum_{i=0}^{N-1} B_{i,k}(t) \mathbf{P}_i, }
\end{equation}
\extension{where \( \mathbf{P}_i \) are learnable control points and \( B_{i,k}(t) \) are spline basis functions defined recursively~\cite{luo2024continuous}. To enhance temporal consistency, the model employs a Neural Ordinary Differential Equation (NODE) module in combination with ConvGRU, where the hidden feature state \( h(t) \) evolves according to:}
\begin{equation}
    \extension{
    \frac{d h(t)}{dt} = f(h(t), t), \quad h(t) = h(t_0) + \int_{t_0}^{t} f(h(\tau), \tau)\, d\tau. }
\end{equation}
\extension{The hidden state \( h(t) \) is refined with frame-specific features \( \varepsilon_t \) via ConvGRU: \( \tilde{h}(t) = \text{ConvGRU}(h(t), \varepsilon_t) \). The model then computes multi-scale correlation volumes between the reference feature \( \tilde{h}(0) \) and \( \tilde{h}(t) \), denoted by \( C(t) = \text{Corr}(\tilde{h}(0), \tilde{h}(t)) \). These correlations are used in an iterative decoder to update the control points of the spline as:}
\begin{equation}
    \sma{
    \mathbf{P}_i^{(s+1)} = \mathbf{P}_i^{(s)} + \Delta \mathbf{P}_i^{(s)}(C(t)),}
\end{equation}
\extension{where \( s \) is the iteration index. The final continuous optical flow $\mathbf{V}_{0 \rightarrow t}$ is reconstructed from the refined control points using the spline formulation above. This continuous formulation allows the model to produce high-fidelity indexing maps using \cref{eq:dt} for arbitrary interpolation times $D_t = \left\Vert \mathbf{V}_{0\to t}\right\Vert \cos{\theta}/\left\Vert \mathbf{V}_{0\to 1} \right\Vert$.}

\subsection{\extension{Multi-frame refiner}}
\sma{We further design a multi-frame fusion module to leverage the relevant pixel appearance information beyond the two adjacent frames. As a brief review, flow-based VFI models~\cite{huang2022real} typically adhere to the following formulation:}
\begin{equation}
    \extension{
    I_{t}^{+}, I_{t}^{-}, M = \mathcal{F}(I_0, I_1, D_t),}
\end{equation}
\extension{where $M$ is a one-channel blending mask, and $I_{t}^{+}$ and $I_{t}^{-}$ denote the warped images derived from $I_0$ and $I_1$ using the predicted optical flows. The final interpolated frame $I_t$ is obtained by:} 
\begin{equation}
    \extension{
    I_t = M \odot I_{t}^{+} + (1-M) \odot I_{t}^{-} }
\end{equation}
\sma{An additional residual term is often included which is omitted here. To adapt the model to accept four consecutive frames as input while maintaining plug-and-play compatibility with various VFI models, we introduce a simple yet effective framework: we create a trainable copy of the two-frame VFI model $\mathcal{F'}$, which accepts the additional frames $I_{-1}$ and $I_2$ along with a new distance map $D_t'$ computed relative to $I_{-1}$ and $I_2$. Additionally, the initially interpolated frame $I_t$ is provided, enabling the network to refine the result by utilizing this supplementary information:}
\begin{equation}
    \extension{
    I_{t}^{'+}, I_{t}^{'-}, M'= \mathcal{F'}(I_{-1}, I_2, D'_t, I_t), }
\end{equation}
\extension{where \sma{$I_{t}^{'+}, I_{t}^{'-}$ are warped version of $I_{-1}, I_2$ respectively.} \( M' = [M_1, M_2, M_3] \) is a three-channel blending mask such that \( M_1 + M_2 + M_3 = 1 \) at each pixel, corresponding to \( I_{t}^{'+} \), \( I_t \), and \( I_{t}^{'-} \). The final refined frame \( I'_t \) is then computed as:}
\begin{equation}
    \extension{
    I'_t = M_1 \odot I_{t}^{'+} + M_2 \odot I_t + M_3 \odot I_{t}^{'-} }
\end{equation}
\sma{Notice that we directly use the interpolated frame $I_t$ instead of latent features as network input to ensure compatibility with different VFI models. }

\begin{figure*}[htbp]
    \begin{center}
        \includegraphics[width=\linewidth]{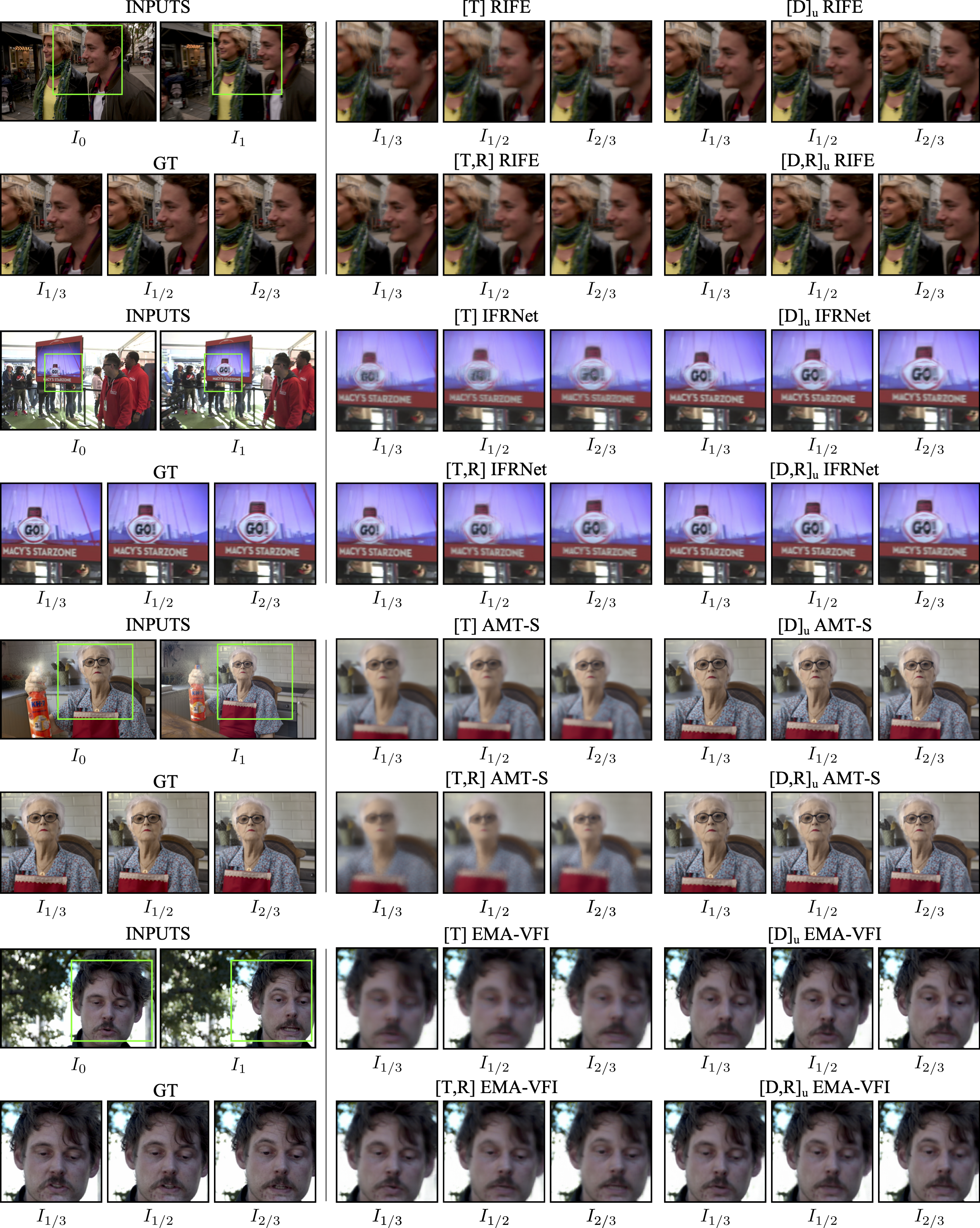}
    \end{center}
\caption{\extension{Comparison of qualitative results. \textbf{$[T]$}: original arbitrary time VFI models using time indexing. \textbf{$[D]_u$}: models trained using our distance indexing, then inference using uniform maps. \textbf{$[T,R]$}: models using time indexing with iterative reference-based estimation. \textbf{$[D,R]_u$}: models trained using both distance indexing and iterative reference-based estimation, then inference using uniform maps.}}
\label{fig:comparison}
\end{figure*}

\begin{figure*}[htbp]
    \begin{center}
        \includegraphics[width=\linewidth]{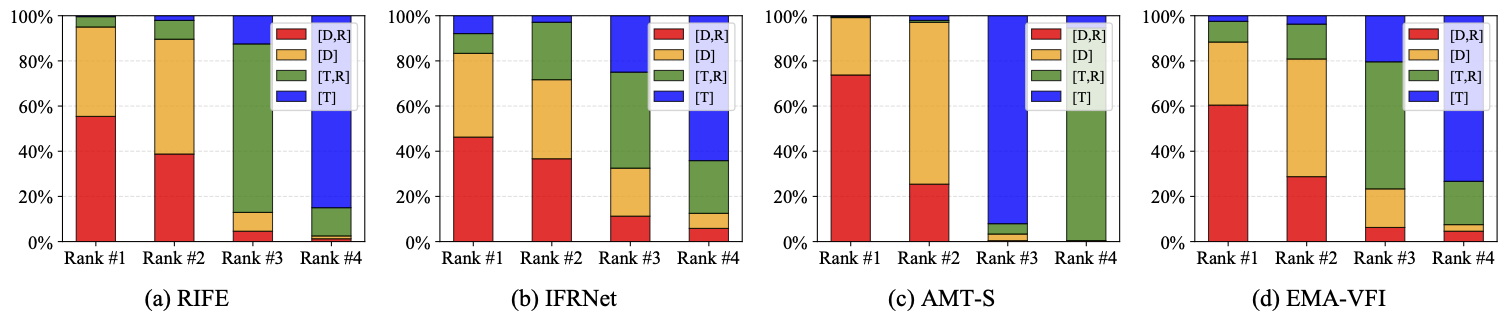}
    \end{center}
    \caption{User study. The horizontal axis represents user rankings, where \#1 is the best and \#4 is the worst. The vertical axis indicates the percentage of times each model variant received a specific ranking. Each model variant was ranked an equal number of times. The model $[D,R]$ emerged as the top performer. All models use uniform maps.}
    \label{fig:user_study}
\end{figure*}

\subsection{\extension{Model tuning}}
\sma{During training, we first freeze the parameters of the two-frame VFI model $\mathcal{F}$ and the continuous map estimator. Only the refiner $\mathcal{F'}$ is updated during this stage, using the same training configuration as the original model. This setup enables the refiner to learn how to enhance interpolation by leveraging information from more distant adjacent frames. Next, we freeze the parameters of the pretrained map estimator and jointly optimize both the VFI model $\mathcal{F}$ and the proposed refiner $\mathcal{F'}$. This approach enables the entire system to adapt to indexing maps derived from optical flow predicted by CPFlow. We experimented with various training strategies and loss functions to evaluate their effectiveness (see \cref{ablation:tuning}).}

\section{Experiment}
\label{sec:experiment}
\subsection{Implementation}
We leveraged the plug-and-play nature of our distance indexing and iterative reference-based estimation strategies to seamlessly integrate them into influential arbitrary time VFI models such as RIFE~\cite{huang2022real} and IFRNet~\cite{kong2022ifrnet}, and state-of-the-art models including AMT~\cite{li2023amt} and EMA-VFI~\cite{zhang2023extracting}. 
\extension{To further strengthen the empirical evidence, we additionally validated the effectiveness of distance indexing on a diffusion-based model, LDMVFI~\cite{danier2024ldmvfi}, and a representative multi-frame method, VFI-Transformer~\cite{shi2022video}.}
We adhere to the original hyperparameters for each model for a fair comparison and implement them with PyTorch~\cite{paszke2019pytorch}.
For training, we use the septuplet dataset from Vimeo90K~\cite{xue2019video}. The septuplet dataset comprises 91,701 seven-frame sequences at $448 \times 256$, extracted from 39,000 video clips. For evaluation, we use both pixel-centric metrics like PSNR and SSIM~\cite{wang2004image}, and perceptual metrics such as reference-based LPIPS~\cite{zhang2018unreasonable} and non-reference NIQE~\cite{mittal2012making}.
Concerning the iterative reference-based estimation strategy, $D_{ref}$ during training is calculated from the optical flow derived from ground-truth data at a time point corresponding to a randomly selected reference frame, like $t/2$. In the inference phase, we similarly employ a uniform map for reference, for example, setting $D_{ref}=t/2$. 

\subsection{Qualitative comparison}
\subsubsection{\extension{Qualitative analysis}} We conducted a qualitative analysis on different variants of each arbitrary time VFI model. We evaluate the base model, labeled as $[T]$, against its enhanced versions, which incorporate distance indexing ($[D]$), iterative reference-based estimation ($[T,R]$), or a combination of both ($[D,R]$), as shown in \cref{fig:comparison}. 
We observe that the $[T]$ model yields blurry results  with details difficult to distinguish. Models with the distance indexing ($[D]$) mark a noticeable enhancement in perceptual quality, presenting clearer interpolations than $[T]$. 
In most cases, iterative reference ($[T,R]$) also enhances model performance, with the exception of AMT-S. As expected, the combined approach $[D,R]$ offers the best quality for all base models including AMT-S. This highlights the synergistic potential of distance indexing when paired with iterative reference-based estimation. Overall, our findings underscore the effectiveness of both techniques as plug-and-play strategies, capable of significantly elevating the qualitative performance of cutting-edge arbitrary time VFI models.

\subsubsection{\extension{User study}} \extension{To validate the effectiveness of our proposed strategies, we further conducted a user study with 30 anonymous participants. Participants were tasked with ranking the interpolation quality of frames produced by four model variants: $[T]$, $[D]$, $[T,R]$, and $[D,R]$. See details of user study UI in Appendix. The results, presented in \cref{fig:user_study}, align with our qualitative and quantitative findings. The $[D,R]$ model variant emerged as the top-rated, underscoring the effectiveness of our strategies.}

\begin{figure*}[htbp]
    \begin{center}
    \includegraphics[width=\linewidth]{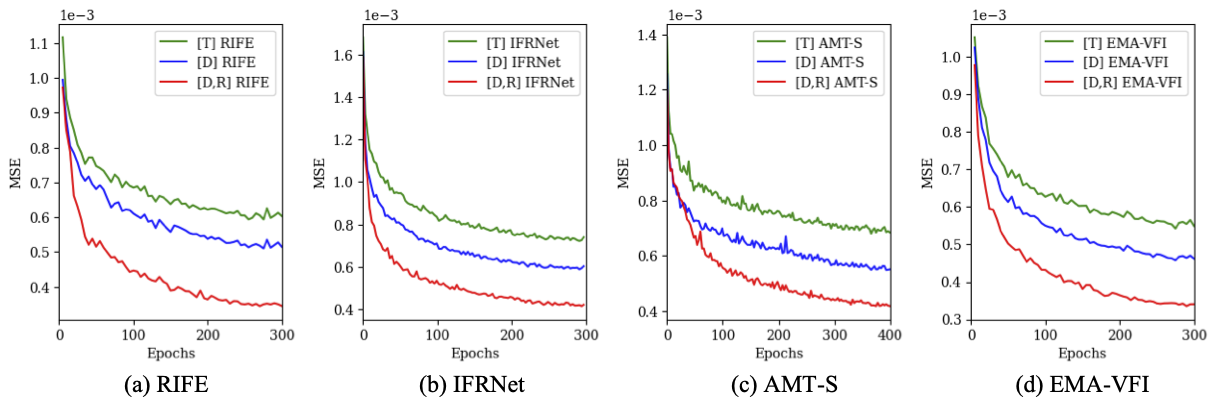}
    \end{center}
\caption{Convergence curves. $[T]$ denotes traditional time indexing. $[D]$ denotes the proposed distance indexing. $[R]$ denotes iterative reference-based estimation.}
\label{fig:convergence}
\end{figure*}

\begin{table*}[!t]
\caption{Comparison on Vimeo90K septuplet dataset. $[T]$ denotes the method trained with traditional arbitrary time indexing paradigm. $[D]$ and $[R]$ denote the distance indexing paradigm and iterative reference-based estimation strategy, respectively. $[R]$ uses 2 iterations by default. $[\cdot]_u$ denotes inference with uniform map as time indexes. We utilize the first and last frames as inputs to predict the rest five frames. The \bestscore{bold font} denotes the best performance in cases where comparison is possible. While the \fakescore{gray font} indicates that the scores for pixel-centric metrics, PSNR and SSIM, are not calculated using strictly aligned ground-truth and predicted frames.
}
\label{tab:vimeo90k_septuplet}
\setlength{\tabcolsep}{8pt}
\begin{center}
  \begin{tabular}{l|ccc|ccc|ccc|ccc|}
    \toprule
    & \multicolumn{3}{c|}{RIFE~\cite{huang2022real}} & \multicolumn{3}{c|}{IFRNet~\cite{kong2022ifrnet}} & \multicolumn{3}{c|}{AMT-S~\cite{li2023amt}} & \multicolumn{3}{c|}{EMA-VFI~\cite{zhang2023extracting}} \\
    \cmidrule(r){2-4}
    \cmidrule(r){5-7} 
    \cmidrule(r){8-10}
    \cmidrule(r){11-13}
    & $[T]$ & $[D]$ & $[D,R]$ & $[T]$ & $[D]$ & $[D,R]$ & $[T]$ & $[D]$ & $[D,R]$ & $[T]$ & $[D]$ & $[D,R]$\\ 
    \midrule
    PSNR$\uparrow$ & 28.22 & \bestscore{29.20} & 28.84 & 28.26 & \bestscore{29.25} & 28.55 & 28.52 & \bestscore{29.61} & 28.91 & 29.41 & \bestscore{30.29} & 25.10 \\
    SSIM$\uparrow$ & 0.912 & \bestscore{0.929} & 0.926 & 0.915 & \bestscore{0.931} & 0.925 & 0.920 & \bestscore{0.937} & 0.931 & 0.928 & \bestscore{0.942} & 0.858 \\
    LPIPS$\downarrow$ & 0.105 & 0.092 & \bestscore{0.081} & 0.088 & 0.080 & \bestscore{0.072} & 0.101 & 0.086 & 
    \bestscore{0.077} & 0.086 & \bestscore{0.078} & 0.079 \\
    NIQE$\downarrow$ & 6.663 & 6.475 & \bestscore{6.286} & 6.422 & 6.342 & \bestscore{6.241} & 6.866 & 6.656 & \bestscore{6.464} & 6.736 & 6.545 & \bestscore{6.241} \\
    \midrule
    & $[T]$ & $[D]_{u}$ & $[D,R]_{u}$ & $[T]$ & $[D]_{u}$ & $[D,R]_{u}$ & $[T]$ & $[D]_{u}$ & $[D,R]_{u}$ & $[T]$ & $[D]_{u}$ & $[D,R]_{u}$\\
    \midrule
    PSNR$\uparrow$ & \bestscore{28.22} & \fakescore{27.55} & \fakescore{27.41} & \bestscore{28.26} & \fakescore{27.40} & \fakescore{27.13} & \bestscore{28.52} & \fakescore{27.33} & \fakescore{27.17} & \bestscore{29.41} & \fakescore{28.24} & \fakescore{24.73} \\
    SSIM$\uparrow$ & \bestscore{0.912} & \fakescore{0.902} & \fakescore{0.901} & \bestscore{0.915} & \fakescore{0.902} & \fakescore{0.899} & \bestscore{0.920} & \fakescore{0.902} & \fakescore{0.902} & \bestscore{0.928} & \fakescore{0.912} & \fakescore{0.851} \\
    LPIPS$\downarrow$ & 0.105 & 0.092 & \bestscore{0.086} & 0.088 & 0.083 & \bestscore{0.078} & 0.101 & 0.090 & \bestscore{0.081} & 0.086 & \bestscore{0.079} & 0.081 \\
    NIQE$\downarrow$ & 6.663 & 6.344 & \bestscore{6.220} & 6.422 & 6.196 & \bestscore{6.167} & 6.866 & 6.452 & \bestscore{6.326} & 6.736 & 6.457 & \bestscore{6.227} \\
    \bottomrule
  \end{tabular}
\end{center}
\end{table*}

\subsection{Quantitative comparison}
\subsubsection{Convergence curves} To further substantiate the efficacy of our proposed strategies, we also conducted a quantitative analysis. \cref{fig:convergence} shows the convergence curves for different model variants, \ie, $[T]$, $[D]$, and $[D,R]$. The observed trends are consistent with our theoretical analysis from \cref{sec:disambiguate}, supporting the premise that by addressing velocity ambiguity, both distance indexing and iterative reference-based estimation can enhance convergence limits.

\subsubsection{Comparison on Vimeo90K septuplet dataset} In \cref{tab:vimeo90k_septuplet}, we provide a performance breakdown for each model variant. The models $[D]$ and $[D,R]$ in the upper half utilize ground-truth distance guidance, which is not available at inference in practice. The goal here is just to show the achievable upper-bound performance. On both pixel-centric metrics such as PSNR and SSIM, and perceptual measures like LPIPS and NIQE, the improved versions $[D]$ and $[D,R]$ outperform the base model $[T]$. Notably, the combined model $[D,R]$ using both distance indexing and iterative reference-based estimation strategies performs superior in perceptual metrics, particularly NIQE. The superior pixel-centric scores of model $[D]$ compared to model $[D,R]$ can be attributed to the indirect estimation (2 iterations) in the latter, causing slight misalignment with the ground-truth, albeit with enhanced details. 

\begin{table*}[htbp]
\caption{Ablation study of the number of iterations on Vimeo90K septuplet dataset. $[\cdot]^{\#}$ denotes the number of iterations used for inference.}
\label{tab:ablation_iterations}
\setlength{\tabcolsep}{8pt}
\begin{center}
  \begin{tabular}{l|ccc|ccc|ccc|ccc|}
    \toprule
    & \multicolumn{3}{c|}{RIFE~\cite{huang2022real}} & \multicolumn{3}{c|}{IFRNet~\cite{kong2022ifrnet}} & \multicolumn{3}{c|}{AMT-S~\cite{li2023amt}} & \multicolumn{3}{c|}{EMA-VFI~\cite{zhang2023extracting}} \\
    \midrule
    $[D,R]_{u}$ & $[\cdot]^{1}$ & $[\cdot]^{2}$ & $[\cdot]^{3}$ & $[\cdot]^{1}$ & $[\cdot]^{2}$ & $[\cdot]^{3}$ & $[\cdot]^{1}$ & $[\cdot]^{2}$ & $[\cdot]^{3}$ & $[\cdot]^{1}$ & $[\cdot]^{2}$ & $[\cdot]^{3}$\\
    \midrule
    LPIPS$\downarrow$ & 0.093 & 0.086 & \bestscore{0.085} & 0.085 & \bestscore{0.078} & 0.078 & 0.086 & \bestscore{0.081} & 0.081 & 0.084 & 0.081 & \bestscore{0.080}\\
    NIQE$\downarrow$ & 6.331 & 6.220 & \bestscore{6.186} & 6.205 & 6.167 & \bestscore{6.167} & 6.402 & \bestscore{6.326} & 6.327 & 6.303 & 6.227 & \bestscore{6.211}\\
    \midrule
    $[T,R]$ & $[\cdot]^{1}$ & $[\cdot]^{2}$ & $[\cdot]^{3}$ & $[\cdot]^{1}$ & $[\cdot]^{2}$ & $[\cdot]^{3}$ & $[\cdot]^{1}$ & $[\cdot]^{2}$ & $[\cdot]^{3}$ & $[\cdot]^{1}$ & $[\cdot]^{2}$ & $[\cdot]^{3}$\\
    \midrule
    LPIPS$\downarrow$ & 0.103 & 0.087 & \bestscore{0.087} & 0.091 & 0.084 & \bestscore{0.084} & \bestscore{0.106} & 0.135 & 0.157 & 0.088 & \bestscore{0.083} & 0.085 \\
    NIQE$\downarrow$ & 6.551 & 6.300 & \bestscore{6.206} & 6.424 & 6.347 & \bestscore{6.314} & \bestscore{6.929} & 7.246 & 7.502 & 6.404 & 6.280 & \bestscore{6.246} \\
    \bottomrule
  \end{tabular}
\end{center}
\end{table*}

In realistic scenarios where the precise distance map is inaccessible at inference, one could resort to a uniform map akin to time indexing. The bottom segment of \cref{tab:vimeo90k_septuplet} shows the performance of the enhanced models $[D]$ and $[D,R]$, utilizing identical inputs as model $[T]$. Given the misalignment between predicted frames using a uniform distance map and the ground-truth, the enhanced models do not outperform the base model on pixel-centric metrics. However, we argue that in most applications, the goal of VFI is not to predict pixel-wise aligned frames, but to generate plausible frames with high perceptual quality. Furthermore, pixel-centric metrics are less sensitive to blur~\cite{zhang2018unreasonable}, the major artifact introduced by velocity ambiguity. The pixel-centric metrics are thus less informative and denoted in gray. On perceptual metrics (especially NIQE), the enhanced models significantly outperforms the base model. This consistency with our qualitative observations further validates the effectiveness of distance indexing and iterative reference-based estimation.

\subsubsection{Ablation study of the number of iterations} \cref{tab:ablation_iterations} offers an ablation study on the number of iterations and the efficacy of a pure iterative reference-based estimation strategy. The upper section suggests that setting iterations at two strikes a good trade-off between computational efficiency and performance. The lower segment illustrates that while iterative reference-based estimation generally works for time indexing, there are exceptions, as observed with AMT-S. However, when combined with distance indexing, iterative reference-based estimation exhibits more stable improvement, as evidenced by the results for $[D,R]_u$. This is consistent with qualitative comparison. 

\begin{table}[htbp]
    \centering
    \caption{Comparison on X4K1000FPS~\cite{sim2021xvfi} for $\times$16 interpolation with RIFE~\cite{huang2022real}.}
    \label{tab:x4k1000fps_x16}
    \setlength{\tabcolsep}{12pt}
    \begin{tabular}{lccc}
        \toprule
        & $[T]$ & $[D]_{u}$ & $[D,R]_{u}$ \\
        \midrule
        PSNR $\uparrow$ & 31.04 & \bestscore{\fakescore{31.60}} & \fakescore{31.52} \\
        SSIM $\uparrow$ & 0.910 & \fakescore{0.914} & \bestscore{\fakescore{0.922}} \\
        LPIPS $\downarrow$ & 0.104 & 0.094 & \bestscore{0.079} \\
        NIQE $\downarrow$ & 7.215 & 6.953 & \bestscore{6.927} \\
        \bottomrule
    \end{tabular}
\end{table}

\subsubsection{Comparison on other benchmarks} The septuplet set of Vimeo90K~\cite{xue2019video} is large enough to train a practical video frame interpolation model, and it represents the situations where the temporal distance between input frames is large. Thus, Vimeo90K (septuplet) can well demonstrate the velocity ambiguity problem that our work aims to highlight. We further show $\times$16 interpolation on X4K1000FPS with larger temporal distance in \cref{tab:x4k1000fps_x16}. The results highlight that the benefits of our strategies are more pronounced with increased temporal distances. 

\begin{table}[htbp]
    \centering
    \caption{Comparison on Vimeo90K~\cite{xue2019video} using GMFlow~\cite{xu2022gmflow} for distance map calculation with RIFE~\cite{huang2022real}.}
    \label{tab:gmflow}
    \setlength{\tabcolsep}{12pt}
    \begin{tabular}{lccc}
        \toprule
        & $[T]$ & $[D]_{u}$ & $[D,R]_{u}$ \\
        \midrule
        PSNR $\uparrow$ & \bestscore{28.22} & \fakescore{27.29} & \fakescore{26.96} \\
        SSIM $\uparrow$ & \bestscore{0.912} & \fakescore{0.898} & \fakescore{0.895} \\
        LPIPS $\downarrow$ & 0.105 & 0.101 & \bestscore{0.092} \\
        NIQE $\downarrow$ & 6.663 & 6.449 & \bestscore{6.280} \\
        \bottomrule
    \end{tabular}
\end{table}

\subsubsection{Other optical flow estimator} We also employ GMFlow~\cite{xu2022gmflow} for the precomputation of distance maps, enabling an analysis of model performance when integrated with alternative optical flow estimations. The results are as shown in \cref{tab:gmflow}. Our strategies still lead to consistent improvement on perceptual metrics. However, this more recent and performant optical flow estimator does not introduce improvement compared to RAFT~\cite{teed2020raft}. A likely explanation is that since we quantify the optical flow to $[0, 1]$ scalar values for better generalization, our training strategies are less sensitive to the precision of the optical flow estimator.

\begin{table}[htbp]
\caption{\extension{Comparison on the septuplet of Vimeo90K \cite{xue2019video} using LPIPS loss \cite{zhang2018unreasonable}. We use RIFE \cite{huang2022real} as a representative example.}}
\label{tab:lpips}
\setlength{\tabcolsep}{12pt}
\begin{center}
  \begin{tabular}{lccc}
    \toprule
    & $[T]$ & $[D]_{u}$ & $[D,R]_{u}$ \\
    \midrule
    PSNR $\uparrow$ & \bestscore{27.19} & \fakescore{26.71} & \fakescore{26.72} \\
    SSIM $\uparrow$ & \bestscore{0.898} & \fakescore{0.889} & \fakescore{0.890} \\
    LPIPS $\downarrow$ & \bestscore{0.061} & 0.065 & 0.064 \\
    NIQE $\downarrow$ & 6.307 & 5.901 & \bestscore{5.837} \\
    \bottomrule
  \end{tabular}
\end{center}
\end{table}

\subsubsection{\extension{Comparison of using perceptual loss}}
\extension{In addition to training with traditional pixel losses based on L1 and L2 losses, we present the results of employing the more recent LPIPS loss \cite{zhang2018unreasonable} with a VGG backbone \cite{simonyan2014very}, as shown in \cref{tab:lpips}. The non-reference perceptual quality metric, NIQE, shows notable improvement across all variants. The results also consistently demonstrate the effectiveness of our strategies in resolving velocity ambiguity. Besides, due to the direct optimization of LPIPS loss, the assumption of constant speed in uniform maps affects the performance for this metric. This is why in the test results, $[T]$ has a lower LPIPS, while $[D]_u$ and $[D,R]_u$ are slightly higher.}

\begin{table*}[htbp]
\caption{\extension{Temporal consistency evaluation on Vimeo90K Septuplet using FVD~\cite{unterthiner2018towards} and VBench~\cite{huang2024vbench}.}}
\label{tab:temporal_metrics}
\setlength{\tabcolsep}{8pt}
\begin{center}
  \begin{tabular}{l|cc|cc|cc|cc|}
    \toprule
    & \multicolumn{2}{c|}{RIFE~\cite{huang2022real}} & \multicolumn{2}{c|}{IFRNet~\cite{kong2022ifrnet}} & \multicolumn{2}{c|}{AMT-S~\cite{li2023amt}} & \multicolumn{2}{c|}{EMA-VFI~\cite{zhang2023extracting}} \\
    \cmidrule(r){2-3}
    \cmidrule(r){4-5} 
    \cmidrule(r){6-7}
    \cmidrule(r){8-9}
    & $[T]$  & $[D]_u$ & $[T]$ & $[D]_u$ & $[T]$  & $[D]_u$ & $[T]$  & $[D]_u$\\  
    \midrule
    FVD $\downarrow$ & 0.0174 & \bestscore{0.0137} & 0.0142 & \bestscore{0.0137} & 0.0167 & \bestscore{0.0137} & 0.0119 & \bestscore{0.0108} \\
    Subject Consistency $\uparrow$ & 0.957 & \bestscore{0.959} & 0.958 & \bestscore{0.961} & 0.958 & \bestscore{0.960} & 0.962 & \bestscore{0.967}  \\
    Background Consistency $\uparrow$ & 0.952 &  \bestscore{0.955} & 0.952 & \bestscore{0.956} & 0.953 & \bestscore{0.955} & 0.955 & \bestscore{0.958} \\
    Imaging Quality $\uparrow$ & 0.448 & \bestscore{0.475} & 0.460 & \bestscore{0.481} & 0.462 & \bestscore{0.485} & 0.468 & \bestscore{0.486} \\
    Motion Smoothness $\uparrow$ & \bestscore{0.996} & \bestscore{0.996} & \bestscore{0.996} & \bestscore{0.996} & \bestscore{0.996} & \bestscore{0.996} & \bestscore{0.996} & \bestscore{0.996}
    \\
    \bottomrule
  \end{tabular}
\end{center}
\end{table*}

\subsubsection{\extension{Temporal consistency}} 
\extension{We have further evaluated models using FVD~\cite{unterthiner2018towards} and VBench~\cite{huang2024vbench}. As shown in~\cref{tab:temporal_metrics}, replacing time indexing with distance indexing consistently improves FVD scores as well as VBench metrics related to subject consistency, background consistency, and overall imaging quality. We notice that the \emph{motion smoothness} metric in VBench~\cite{huang2024vbench} shows limited discriminative power for video frame interpolation. These results indicate that reducing velocity ambiguity not only enhances per-frame perceptual quality but also leads to more temporally coherent and stable video interpolation.}

\begin{figure*}[htbp]
    \begin{center}
    \includegraphics[width=\linewidth]{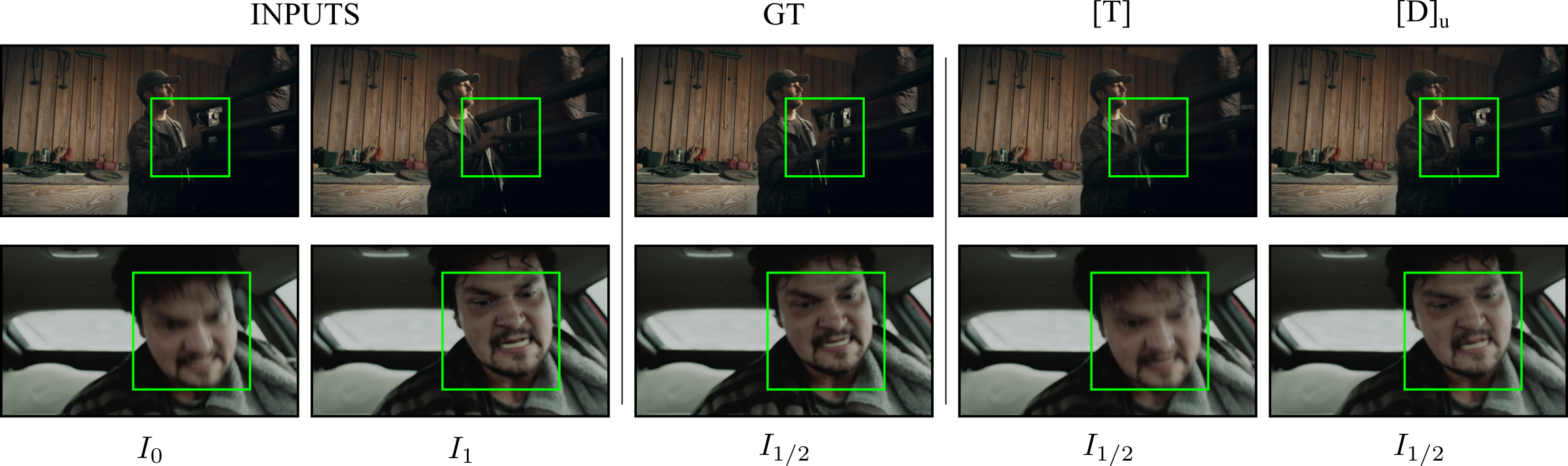}
    \end{center}
    \caption{\extension{Comparison on Vimeo90K Septuplet with LDMVFI~\cite{danier2024ldmvfi}.}}
    \label{fig:ldmvfi}
\end{figure*}

\begin{table}[htbp]
\caption{\extension{Comparison on Vimeo90K Septuplet with LDMVFI~\cite{danier2024ldmvfi}.}}
\label{tab:vimeo90k_ldmvfi}
\setlength{\tabcolsep}{8pt}
\centering
\begin{tabular}{l|ccc}
\toprule
 & $[T]$ & $[D]$ & $[D]_u$ \\
\midrule
PSNR$\uparrow$  & 26.83 & \bestscore{27.12} & \fakescore{25.94} \\
SSIM$\uparrow$  & 0.904 & \bestscore{0.906} & \fakescore{0.893} \\
LPIPS$\downarrow$ & 0.098 & \bestscore{0.086} & 0.090 \\
NIQE$\downarrow$  & 6.652 & 6.481 & \bestscore{6.451} \\
\bottomrule
\end{tabular}
\end{table}

\subsection{\extension{Evaluating distance indexing on diffusion-based baseline}}
\extension{
We further evaluate diffusion-based VFI models, which are known for their strong generative capacity. Since SVDKFI~\cite{wang2024generative} directly generates a fixed sequence of multiple frames in a single pass, we adopt LDMVFI~\cite{danier2024ldmvfi} for evaluation, whose formulation is more compatible with our interpolation setting.
As reported in~\cref{tab:vimeo90k_ldmvfi}, even though diffusion models may partially mitigate ambiguity through generative priors, replacing time indexing with distance indexing still leads to further improvements. We also provide a qualitative comparison in~\cref{fig:ldmvfi}. These results indicate that velocity ambiguity remains relevant for diffusion-based approaches and that distance indexing provides complementary benefits.
}

\begin{figure*}[htbp]
    \begin{center}
    \includegraphics[width=\linewidth]{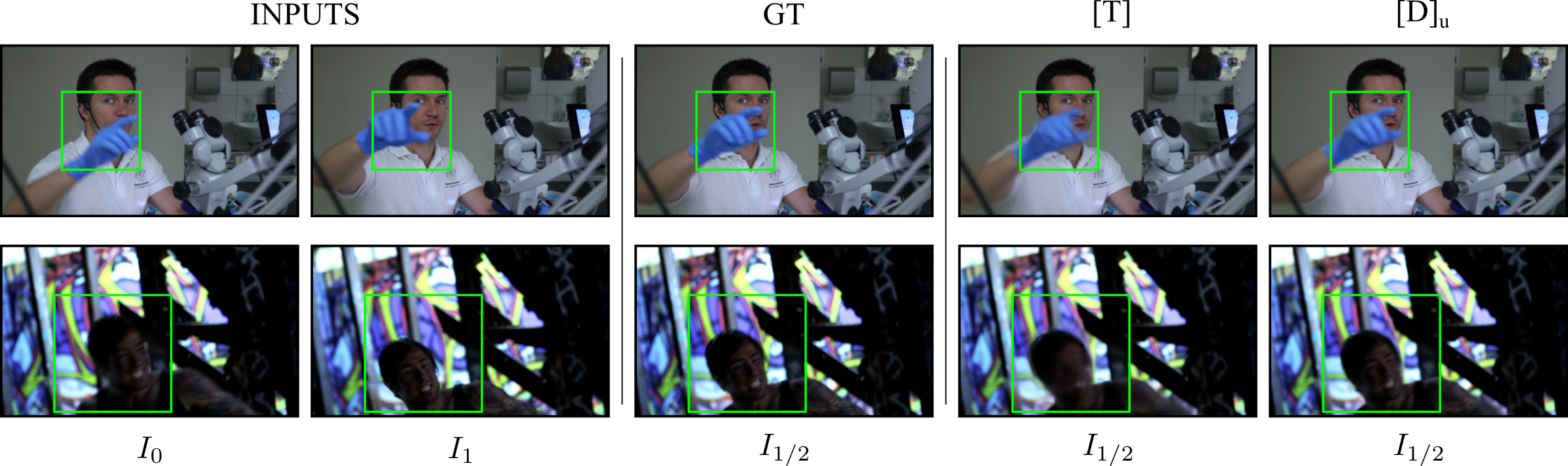}
    \end{center}
    \caption{\extension{Comparison on Vimeo90K Septuplet with VFI-Transformer~\cite{zhang2023extracting}.}}
    \label{fig:vfitransformer}
\end{figure*}

\begin{table}[htbp]
\caption{\extension{Comparison on Vimeo90K Septuplet with VFI-Transformer~\cite{zhang2023extracting}.}}
\label{tab:vimeo90k_vfitransformer}
\setlength{\tabcolsep}{8pt}
\centering
\begin{tabular}{l|ccc}
\toprule
 & $[T]$ & $[D]$ & $[D]_u$ \\
\midrule
PSNR$\uparrow$  & 36.09 & \bestscore{36.46} & \fakescore{35.43} \\
SSIM$\uparrow$  & 0.974 & \bestscore{0.976} & \fakescore{0.964} \\
LPIPS$\downarrow$ & 0.084 & \bestscore{0.079} & 0.081 \\
NIQE$\downarrow$  & 6.251 & \bestscore{6.082} & 6.197 \\
\bottomrule
\end{tabular}
\end{table}

\subsection{\extension{Evaluating distance indexing on multi-frame baseline}}
\extension{
We also evaluate a representative multi-frame input method, Video Frame Interpolation Transformer (VFI-Transformer)~\cite{shi2022video}. As shown in~\cref{tab:vimeo90k_vfitransformer} and~\cref{fig:vfitransformer}, distance indexing also improves performance in the multi-frame setting, confirming that additional temporal context alone does not fully eliminate velocity ambiguity and that the proposed strategy remains beneficial.
}

\subsection{2D manipulation of frame interpolation}
\begin{figure*}[htbp]
    \begin{center}
        \includegraphics[width=\linewidth]{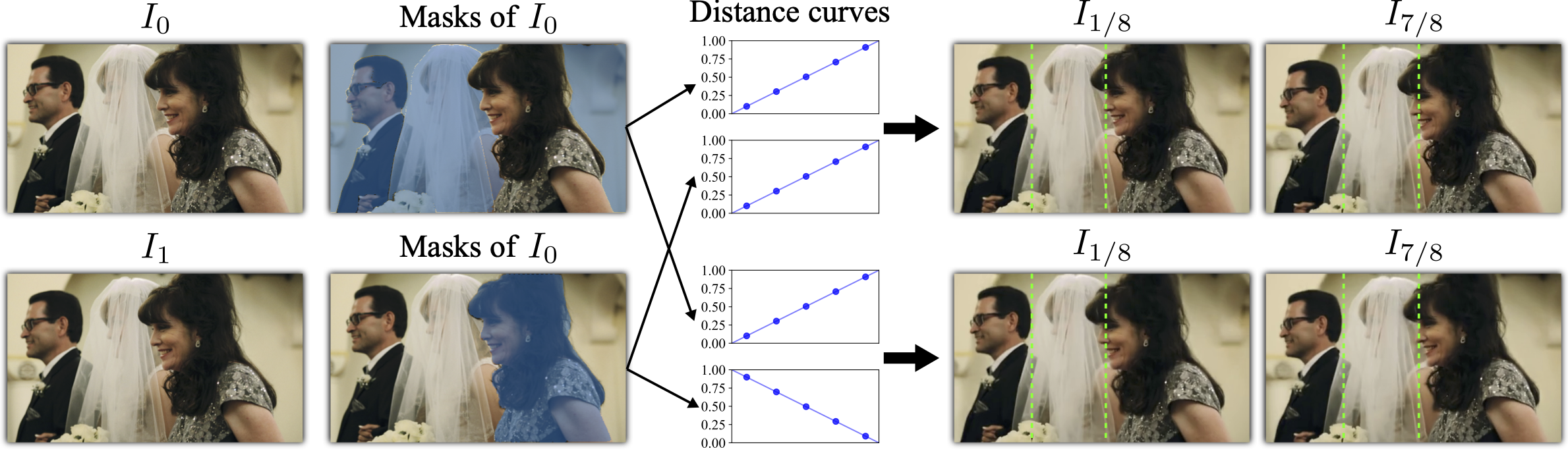}
    \end{center}
    \caption{Manipulated interpolation of anything. Leveraging Segment-Anything~\cite{kirillov2023segment}, users can tailor distance curves for selected masks. Distinct masks combined with varying distance curves generate unique distance map sequences, leading to diverse interpolation outcomes.}
    \label{fig:webapp}
\end{figure*}

Beyond simply enhancing the performance of VFI models, distance indexing equips them with a novel capability: tailoring the interpolation patterns for each individual object, termed as ``manipulated interpolation of anything''. \cref{fig:webapp} demonstrates the workflow. The first stage employs SAM~\cite{kirillov2023segment} to produce object masks for the starting frame. Users can then customize the distance curve for each object delineated by the mask, effectively controlling its interpolation pattern, \eg, having one person moving backward in time. The backend of the application subsequently generates a sequence of distance maps based 
on these specified curves for interpolation. One of the primary applications is re-timing specific objects \textbf{(See the supplementary video)}.

\subsection{\extension{Multi-frame qualitative comparison}}
\begin{figure*}[htbp]
    \begin{center}
        \includegraphics[width=.95\linewidth]{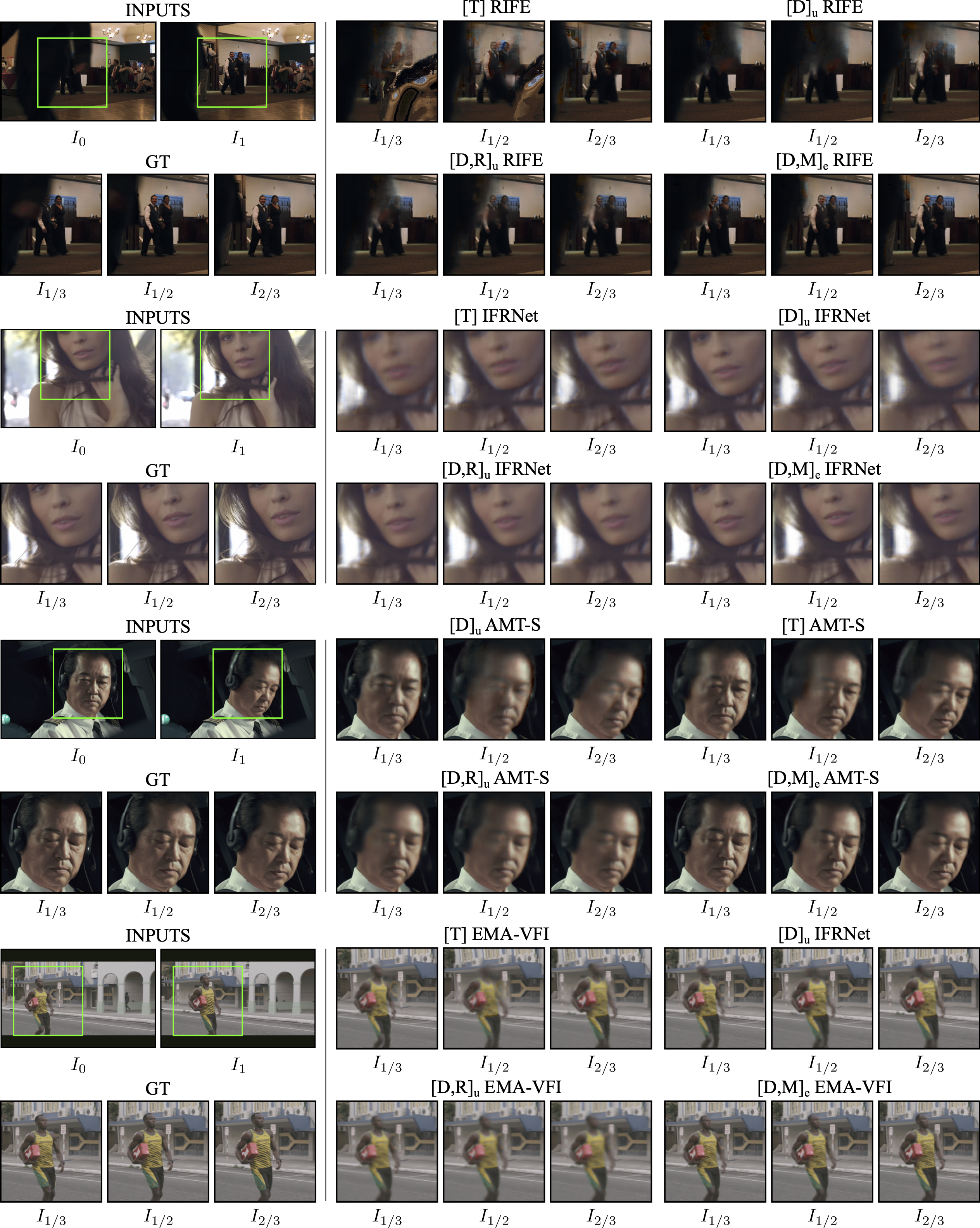}
    \end{center}
\caption{\extension{Qualitative comparison under multi-frame setting. \textbf{$[T]$}: original arbitrary time VFI models using time indexing. \textbf{$[D]_u$}: models trained using distance indexing, then inference using uniform maps. \textbf{$[D,R]_u$}: models trained using both distance indexing and iterative reference-based estimation, then inference using uniform maps. \textbf{$[D,M]_e$}: models trained using distance indexing with multi-frame fusion, then inference using estimated maps. \extension{Due to space limitations, the two adjacent input frames are omitted in the visualization, while the actual input consists of four frames.}}}
\label{fig:multi_qualitative}
\end{figure*}

\extension{As shown in~\cref{fig:multi_qualitative}, we present a qualitative comparison of different variants of each VFI model under multi-frame setup. $[D, M]_e$ denotes the model that incorporate both the indexing map with multi-frame refiner and continuous indexing map estimator. As anticipated, the $[D, M]_e$ configuration consistently yields the highest visual quality across all base models, demonstrating the effectiveness of both the multi-frame refiner and the continuous indexing map estimator.}

\subsection{\extension{Multi-frame quantitative comparison}}
\extension{In~\cref{tab:multi_quantitative}, we present a quantitative performance comparison across various model variants. The configurations $[T]$, $[D]_u$, and $[D]$ retain the same definitions as previously described. $[D]_e$ denotes the model augmented with a continuous indexing map estimator, $[D, M]$ includes the proposed multi-frame fusion with the refiner model, and $[D, M]_e$ incorporates both the refiner and the map estimator.} \extension{$[T, M]$ denotes multi-frame fusion using the refiner model trained with the time indexing map.} \extension{On pixel-level metrics such as PSNR and SSIM, the enhanced variant $[D, M]_e$ consistently outperforms the base model $[T]$ as well as $[D]_u$, which lacks a map estimator and relies on uniform indexing maps. The improvement from $[D]_u$ to $[D, M]_e$ highlights the effectiveness of integrating the continuous indexing map estimator. Meanwhile, performance gains from $[D]$ to $[D, M]$, and from $[D]_e$ to $[D, M]_e$, demonstrate the additional benefit provided by the multi-frame refiner.} \extension{Moreover, the comparison between time-indexed ($[T, M]$) and distance-indexed ($[D,M]_{e}$, $[D,M]$) multi-frame video frame interpolation models again shows that our proposed distance indexing yields better interpolation results.}

\begin{table}[!t]
\setlength{\tabcolsep}{2.5pt}
\caption{\extension{Multi-frame comparison on Vimeo90K septuplet dataset. $[M]$ denotes the multi-frame fusion. $[\cdot]_e$ denotes inference with estimated indexing map. Other notations have the same meaning as in the previous experiments.}}
\begin{center}
  \begin{tabular}{l|ccccc|cc|}
    \toprule
    RIFE~\cite{huang2022real}
    & $[T]$ & $[T, M]$ & $[D]_{u}$ & $[D]_{e}$ & $[D,M]_{e}$ & $[D]$ & $[D,M]$\\
    \midrule
    PSNR $\uparrow$ & 28.22 & \bestscore{28.84} & \fakescore{27.55} & 28.25 & 28.34 & 29.20 & \bestscore{31.63}\\
    SSIM $\uparrow$ & 0.912 & 0.922 & \fakescore{0.902} & 0.923 & \bestscore{0.928} & 0.929 & \bestscore{0.952}\\
    LPIPS $\downarrow$ & 0.105 & 0.097 & 0.092 & 0.099 & \bestscore{0.089} & 0.092 & \bestscore{0.062}\\
    NIQE $\downarrow$ & 6.663 & 6.518 & 6.344 & 6.554 & \bestscore{6.173} & 6.475 & \bestscore{5.990}\\
    \midrule
    IFRNet~\cite{kong2022ifrnet}
    & $[T]$ & $[T, M]$ & $[D]_{u}$ & $[D]_{e}$ & $[D,M]_{e}$ & $[D]$ & $[D,M]$\\
    \midrule
    PSNR $\uparrow$ & 28.26 & \bestscore{28.75} & \fakescore{27.40} & 27.63 & 28.28 & 29.25 & \bestscore{32.11}\\
    SSIM $\uparrow$ & 0.915 & 0.918 & \fakescore{0.902} & 0.907 & \bestscore{0.919} & 0.931 & \bestscore{0.958}\\
    LPIPS $\downarrow$ & 0.088 & 0.085 & 0.083 & 0.087 & \bestscore{0.083} & 0.080 & \bestscore{0.074}\\
    NIQE $\downarrow$ & 6.422 & 6.388 & \bestscore{6.196} & 6.414 & 6.249 & 6.342 & \bestscore{5.957}\\
    \midrule
    AMT-S~\cite{li2023amt}
    & $[T]$ & $[T, M]$ & $[D]_{u}$ & $[D]_{e}$ & $[D,M]_{e}$ & $[D]$ & $[D,M]$\\
    \midrule
    PSNR $\uparrow$ & 28.52 & \bestscore{28.91} & \fakescore{27.33} & 27.60 & 28.80 & 29.61 & \bestscore{31.80}\\
    SSIM $\uparrow$ & 0.920 & \bestscore{0.924} & \fakescore{0.902} & 0.908 & 0.922 & 0.937 & \bestscore{0.955}\\
    LPIPS $\downarrow$ & 0.101 & 0.094 & 0.090 & 0.098 & \bestscore{0.084} & 0.086 & \bestscore{0.072}\\
    NIQE $\downarrow$ & 6.866 & 6.598 & 6.382 & 6.452 & \bestscore{6.223} & 6.656 & \bestscore{6.056}\\
    \midrule
    EMA-VFI~\cite{zhang2023extracting}
    & $[T]$ & $[T, M]$ & $[D]_{u}$ & $[D]_{e}$ & $[D,M]_{e}$ & $[D]$ & $[D,M]$\\
    \midrule
    PSNR $\uparrow$ & 29.41 & \bestscore{29.80} & \fakescore{28.24} & 28.67 & 29.45 & 30.29 & \bestscore{31.31} \\
    SSIM $\uparrow$ & 0.928 & 0.930 & \fakescore{0.912} & 0.919 & \bestscore{0.932
    } & 0.942 & \bestscore{0.951} \\
    LPIPS $\downarrow$ & 0.086 & 0.086 & \bestscore{0.079} & 0.082 & 0.084 & 0.078 & \bestscore{0.069}\\
    NIQE $\downarrow$ & 6.736 & 6.597 & 6.457 & 6.609 & \bestscore{6.313} & 6.545 & \bestscore{6.146}\\
    \bottomrule
  \end{tabular}
\end{center}
\label{tab:multi_quantitative}
\end{table}

\begin{table}[!t]
    \centering
    \begin{minipage}[t]{0.46\textwidth}
        \caption{\extension{Multi-frame ablation study }}
        \setlength{\tabcolsep}{5pt}
        \begin{center}
          \begin{tabular}{lcccccc}
            \toprule
            & \multicolumn{3}{c}{\textit{FE}} & \multicolumn{2}{c}{\textit{VFI}} \\
            \cmidrule(r){2-4}
            \cmidrule(r){5-6}
            & $\mathcal{L}_{D}$ & $\mathcal{L}_{V_{b+f}}$ & $\mathcal{L}_{D+V_{b+f}}$ & $\mathcal{L}_{V_f}$ & $\mathcal{L}_{V_{b+f}}$ \\
            \midrule
            PSNR $\uparrow$ & 26.10 & 27.22 & 27.34 & 30.89 & \bestscore{31.63} \\
            SSIM $\uparrow$ & 0.891 & 0.898 & 0.901 & 0.938 & \bestscore{0.952} \\
            LPIPS $\downarrow$ & 0.111 & 0.104 & 0.097 & 0.069 & \bestscore{0.062} \\
            NIQE $\downarrow$ & 6.627 & 6.575 & 6.509 & 6.087 & \bestscore{5.990} \\
            \bottomrule
          \end{tabular}
        \end{center}
        \label{tab:multi_ablations}
    \end{minipage}
\end{table}

\subsection{\extension{Multi-frame tuning strategy}}
\label{ablation:tuning}
\extension{In this section, we evaluate different training strategies for multi-frame fusion using RIFE~\cite{huang2022real} as the baseline for ablation studies. All experiments follow the same setting of previous experiments on Vimeo90k. We find that jointly tuning both refiner and map estimator leads to instability and failure to converge. Thus, we split the tuning process in two stages. In the first stage, we first finetune the multi-frame refiner $\mathcal{F'}$ alone using original VFI loss. 
\sma{In the second stage, there are two possible choices. Since jointly tuning with the map estimator remains unstable, we either finetune the flow estimator only to adapt it to VFI (referred to as \textit{FE}), or finetune the VFI modules $\mathcal{F}$ and $\mathcal{F'}$ to adapt them to the pretrained flow estimator (referred to as \textit{FE}), as shown in~\cref{tab:multi_ablations}.}
The loss term $\mathcal{L}_D$ is calculated by $\mathcal{L}_D = \| D_t - D_t^{\text{RAFT}} \|_2^2$, denoting the L2 loss between the indexing maps from CPFlow and RAFT. The setting $\mathcal{L}_{V_f}$ corresponds to training only the VFI refiner with its original loss, while $\mathcal{L}_{V_{b+f}}$ involves jointly training both the base VFI model and the refiner with their original loss. Among the tested strategies, jointly tuning both the VFI model and the refiner under $\mathcal{L}_{V_{b+f}}$ achieves the best performance. This improvement is attributed to the model’s ability to adapt both components to the updated indexing map derived from the newly predicted optical flow.}

\begin{table}[!t]
\setlength{\tabcolsep}{4pt}
\caption{\extension{Costs comparison including runtime and number of parameters. We evaluate the computational overhead on an NVIDIA A100 GPU using images with a resolution of 448×256.}}
\begin{center}
  \begin{tabular}{lcccccc}
    \toprule
    & \multicolumn{2}{c}{$[D]$} & \multicolumn{2}{c}{$[D, M]$} & \multicolumn{2}{c}{$[D, M]_e$}\\
    \cmidrule(r){2-3}
    \cmidrule(r){4-5}
    \cmidrule(r){6-7}
    & Sec. & MB & Sec. & MB & Sec. & MB\\
    \midrule
    RIFE~\cite{huang2022real} & 0.03 & 10.21 & 0.06 & 20.46 & 0.10 & 30.68\\
    IFRNet~\cite{kong2022ifrnet} &0.14 & 4.73 & 0.17 & 14.91 & 0.21 & 25.13\\
    AMT-S~\cite{li2023amt} & 0.20 & 2.86 & 0.22 & 15.01 & 0.26 & 25.23\\
    EMA-VFI~\cite{zhang2023extracting} & 2.35 & 62.62 & 2.38 & 74.84 & 2.42 & 85.04\\
    \bottomrule
  \end{tabular}
\end{center}
\label{tab:costs}
\end{table}

\subsection{\sma{Computational costs of the proposed framework}}
\label{sec:cost}
\paragraph{Distance indexing} Transitioning from time indexing ($[T]$) to distance indexing ($[D]$) does not introduce extra computational costs during the inference phase, yet significantly enhancing image quality. In the training phase, the primary requirement is a one-time offline computation of distance maps for image triplets. 

\paragraph{Iterative reference-based estimation} Given that the computational overhead of merely expanding the input channel, while keeping the rest of the structure unchanged, is negligible, the computational burden during the training phase remains equivalent to that of the $[D]$ model. Regarding inference, the total consumption is equal to the number of iterations $\times$ the consumption of the $[D]$ model. We would like to highlight that this iterative strategy is optional: Users can adopt this strategy at will when optimal interpolation results are demanded and the computational budget allows. 

\paragraph{\extension{Continuous indexing map estimation}} 
\extension{The runtime and the number of parameters are reported in~\cref{tab:costs}. The estimation process introduces an additional latency of approximately 0.04 seconds per inference, which is relatively low. The number of parameters for the map estimator is 10.2M. Overall, the computational overheads introduced by indexing map estimation are acceptable in practical scenarios.}

\paragraph{\extension{Multi-frame refiner}} 
\extension{We also evaluate the computational overhead introduced by the multi-frame refiner as reported in~\cref{tab:costs}.  The refiner adds approximately 0.03 seconds per inference and requires an additional 10.3 million parameters, which is practical for deployment.}

\subsection{\extension{Limitations}}
\extension{While the proposed distance indexing strategy effectively stabilizes motion interpolation under large temporal gaps, it does not explicitly address content ambiguity caused by occlusion or severe lighting variations, where visual information is partially missing or altered. The proposed distance indexing relies on distance maps estimated from optical flow. In scenarios involving severe occlusion, curved motion, or significant lighting variations, inaccuracies in optical flow estimation may lead to imperfect distance maps, which can negatively affect the learning process. While our experiments show that distance indexing remains robust to moderate estimation errors, its performance may be limited when such errors become dominant. Nevertheless, we believe that future advances in optical flow estimation will further enhance the effectiveness of distance-indexed video frame interpolation.}

\section{Conclusion}
\label{sec:conclusion}
We challenge the traditional time indexing paradigm and address its inherent uncertainties related to velocity distribution. Through the introduction of distance indexing and iterative reference-based estimation strategies, we offer a transformative paradigm to VFI. Our innovative plug-and-play strategies not only improves the performance in video interpolation but also empowers users with granular control over interpolation patterns across varied objects. 
\extension{We also propose a continuous distance map estimator to accurately predict distance maps when using multi-frame inputs. Additionally, a multi-frame refiner is integrated into the interpolation pipeline for further enhancement. \sma{While the proposed framework significantly improves both pixelwise and perceptual metrics, it still faces challenges in learning and representing diverse interpolation trajectories. Extreme large motions may require generative priors from models like diffusion models.}}
The insights gleaned from our strategies have potential applications across a range of tasks that employ time indexing, such as space-time super-resolution, future predictions, blur interpolation and more.

\section*{Acknowledgments}

We thank Dorian Chan, Zhirong Wu, and Stephen Lin for their insightful feedback and advice. Our thanks also go to Vu An Tran for developing the web application.

{\appendices
\section{Proof of \mainEqAvgPoss}
\label{sec:proof}
\mainEqAvgPoss in the paper can be rigorously proven if an L2 loss is used,
\begin{equation}
    \min_{\hat{I}_t} L=\mathbb{E}_{I_t \sim \mathcal{F}(I_0, I_1, t)}[(\hat{I}_t-I_t)^2].
\end{equation}
By setting the gradient to zero (this assumes during training, the neural network can reach the exact solution at this point),
\begin{equation}
\begin{aligned}
    \frac{\partial L}{\partial \hat{I}_t} &= 0 \\
    \mathbb{E}_{I_t \sim \mathcal{F}(I_0, I_1, t)}[\frac{\partial}{\partial \hat{I}_t}(\hat{I}_t-I_t)^2] &= 0 \\
    \mathbb{E}_{I_t \sim \mathcal{F}(I_0, I_1, t)}[2(\hat{I}_t-I_t)] &= 0 \\
    \mathbb{E}_{I_t \sim \mathcal{F}(I_0, I_1, t)}[\hat{I}_t]-\mathbb{E}_{I_t \sim \mathcal{F}(I_0, I_1, t)}[I_t] &= 0\\
    \hat{I}_t &= \mathbb{E}_{I_t \sim \mathcal{F}(I_0, I_1, t)}[I_t]
\end{aligned}
\end{equation}
\sma{Even when a different loss is used, we still observe a similar phenomenon in practice.}

\section{The Rationale for Solving Ambiguity}
\label{sec:rationale}
First and foremost, it is essential to clarify that \textbf{velocity ambiguity can solely exist and be resolved in the training phase, not in the inference phase}. The key idea behind our approach can be summarized as follows: While conventional VFI methods with time indexing rely on a one-to-many mapping, our distance indexing learns an approximate one-to-one mapping, which resolves the ambiguity during training. When the input-output relationship is one-to-many during training, the training process fluctuates among conflicting objectives, ultimately preventing convergence towards any specific optimization goal. In VFI, the evidence is the generation of blurry images in the inference phase. Once the ambiguity has been resolved using the new indexing method in the training phase, the model can produce significantly clearer results regardless of the inference strategy used.

Indeed, this one-to-many ambiguity in training is not unique to VFI, but for a wide range of machine learning problems. 
It is sometimes referred to as ``mode averaging'' in the community.\footnote{\url{https://www.cs.toronto.edu/~hinton/coursera/lecture13/lec13.pdf}}
In some areas, researchers have come up with similar methods \cite{wang2018style,xu2023dmv3d}.

\subsection*{A specific instantiation of this problem} Let us look at an example in text-to-speech (TTS). The same text can be paired with a variety of speeches, and direct training without addressing ambiguities can result in a \textbf{``blurred'' voice} (a statistical average voice). To mitigate this, a common approach is to incorporate a speaker embedding vector or a style embedding vector (representing different gender, accents, speaking styles, etc.) during training, which helps reduce ambiguity. \textbf{During the inference phase, utilizing an average user embedding vector can yield high-quality speech output.} Furthermore, by manipulating the speaker embedding vector, effects such as altering the accent and pitch can also be achieved.

Here is a snippet from a high-impact paper Wang \etal \cite{wang2018style} which came up with the style embedding in TTS:
\begin{quote}
    Many TTS models, including recent end-to-end systems, only learn an averaged prosodic distribution over their input data, generating less expressive speech – especially for long-form phrases. Furthermore, they often lack the ability to control the expression with which speech is synthesized.
\end{quote}
Understanding this example can significantly help understand our paper, as there are many similarities between the two, \eg, motivation, solution, and manipulation.

\subsection*{A minimal symbolic example to help understand better} Assuming we want to train a mapping function $\mathcal{F}$ from numbers to characters.

\noindent \textbf{Training input-output pairs with ambiguity} ($\mathcal{F}$ is optimized):
\begin{equation*}
    1 \stackrel{\mathcal{F}}{\longrightarrow} a, 1 \stackrel{\mathcal{F}}{\longrightarrow} b, 2 \stackrel{\mathcal{F}}{\longrightarrow} a, 2 \stackrel{\mathcal{F}}{\longrightarrow} b
\end{equation*}
$\mathcal{F}$ is optimized with some losses involving the input-output pairs above:
\begin{equation*}
    \underset{\mathcal{F}}{\min} \quad L(\mathcal{F}(1), a) + L(\mathcal{F}(1),b) + L(\mathcal{F}(2),a) +L(\mathcal{F}(2),b),
\end{equation*}
where $L$ can be L1, L2 or any other kind of losses. Because the same input is paired with multiple different outputs, the model $\mathcal{F}$ is optimized to learn an average (or, generally, a mixture) of the conflicting outputs, which results in blur at inference.

\noindent Inference phase ($\mathcal{F}$ is fixed):
\begin{equation*}
    1 \stackrel{\mathcal{F}}{\longrightarrow} \{a,b\}?, 2 \stackrel{\mathcal{F}}{\longrightarrow} \{a,b\}?
\end{equation*}

\noindent \textbf{Training without ambiguity} ($\mathcal{F}$ is optimized):
\begin{equation*}
    1 \stackrel{\mathcal{F}}{\longrightarrow} a, 1 \stackrel{\mathcal{F}}{\longrightarrow} a, 2 \stackrel{\mathcal{F}}{\longrightarrow} b, 2 \stackrel{\mathcal{F}}{\longrightarrow} b
\end{equation*}
In the input-output pairs above, each input value is paired with exactly one output value. Therefore, $\mathcal{F}$ is trained to learn a unique and deterministic mapping.

\noindent Inference phase ($\mathcal{F}$ is fixed):
\begin{equation*}
    1 \stackrel{\mathcal{F}}{\longrightarrow} a, 2 \stackrel{\mathcal{F}}{\longrightarrow} b
\end{equation*}

\begin{figure*}[htbp]
    \begin{center}
    \includegraphics[width=.85\linewidth]{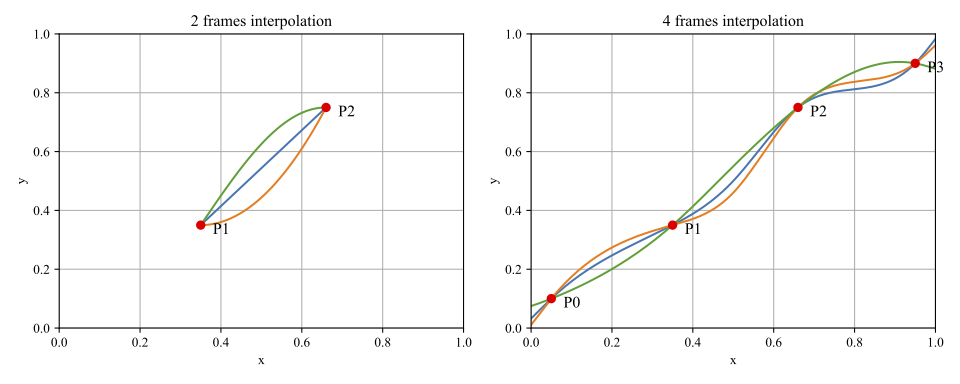}
    \end{center}
    \vspace{-1.5em}
\caption{Analysis of ambiguity for the case of multiple frames.}
\label{fig:multiple_ambiguity}
\end{figure*}

\subsection*{Coming back to VFI}
When time indexing is used, the same $t$ value is paired to images where the objects are located at various locations due to the speed and directional ambiguities. When distance mapping is used, a single $d$ value is paired to images where the objects are always at the same distance ratio, which allows the model to learn a more deterministic mapping for resolving the speed ambiguity.

It is important to note that fixing the ambiguity does not solve all the problems: At inference time, the ``correct'' (close to ground-truth) distance map is not available. In this work, we show that it is possible to provide uniform distance maps as inputs to generate a clear output video, which is not perfectly pixel-wise aligned with the ground truth. This is the reason why the proposed method does not achieve state-of-the-art in terms of PSNR and SSIM in Tab.~1 of the paper. However, it achieves sharper frames with higher perceptual quality, which is shown by the better LPIPS and NIQE.

We claim the ``correct'' distance map is hard to estimate accurately from merely two frames since there are a wide range of possible velocities. If considering more neighboring frames like Xu \etal \cite{xu2019quadratic} (more observation information), it is possible to estimate an accurate distance map for pixel-wise aligned interpolation.

Furthermore, manipulating distance maps corresponds to sampling other possible unseen velocities, \ie, 2D manipulation of frame interpolation, similar to that mentioned TTS paper Wang \etal \cite{wang2018style}.

\subsection*{\extension{The case of multi-frame inputs}}
\extension{Multi-frame inputs introduce additional motion priors, which help constrain possible trajectories but cannot fully resolve velocity ambiguity.
To clarify this, here is a theoretical illustration demonstrating that as shown in \cref{fig:multiple_ambiguity}.
And even when more frames are available, the velocity profile along the trajectory remains under-determined.}

\begin{figure*}[!t]
    \begin{center}
        \includegraphics[width=\linewidth]{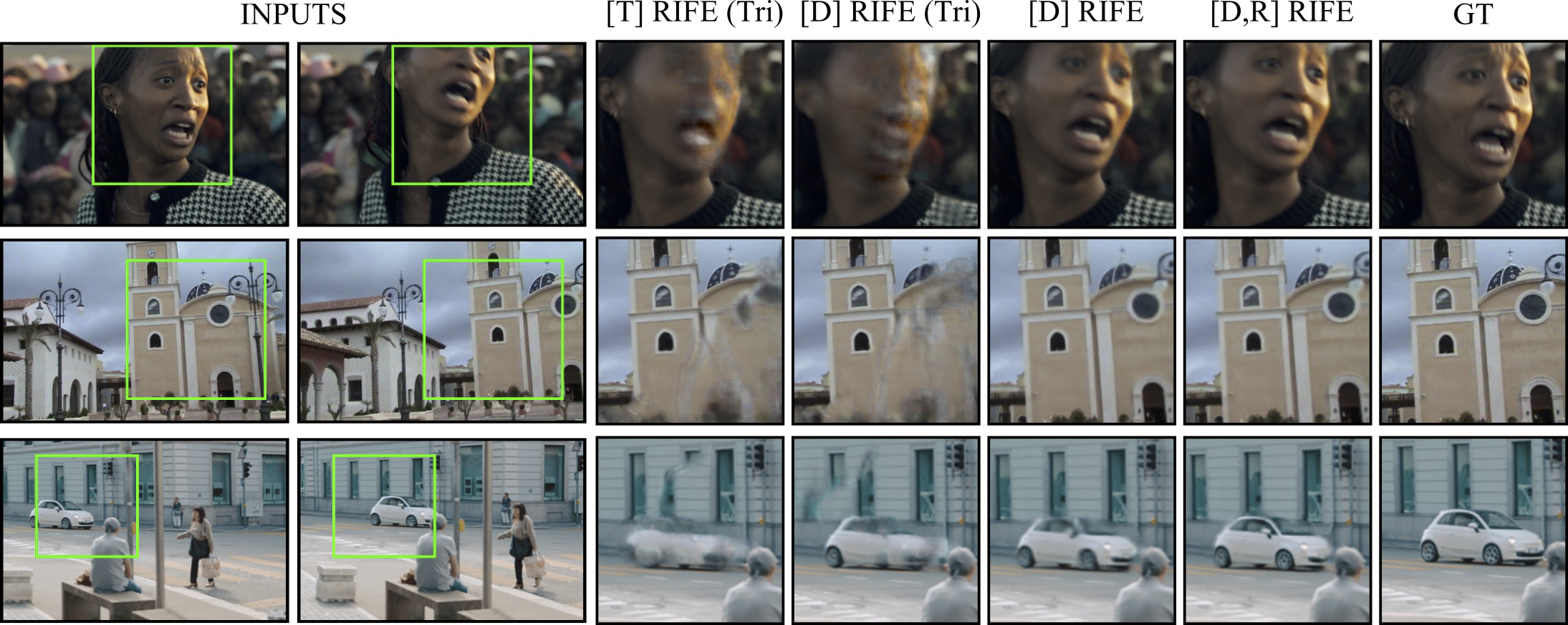}
    \end{center}
\caption{Additional comparison of qualitative results. [T] RIFE (Tri) denotes RIFE \cite{huang2022real} trained in a fixed time indexing paradigm (Vimeo90K triplet dataset \cite{xue2019video}). [D] RIFE (Tri) denotes the model trained using distance indexing. All models use uniform maps.}
\label{fig:additional_qualitative_comparison}
\end{figure*}

\begin{table*}[!t]
\caption{Comparison on Vimeo90K triplet dataset. $[T]$ denotes the method trained with traditional fixed time indexing paradigm. $[D]$ denotes the distance indexing paradigm. $[\cdot]_u$ denotes inference with uniform map as time indexes.}
\label{tab:vimeo90k_triplet}
\setlength{\tabcolsep}{9pt}
\begin{center}
  \begin{tabular}{l|ccc|ccc|ccc|ccc|}
    \toprule
    & \multicolumn{3}{c|}{RIFE \cite{huang2022real}} & \multicolumn{3}{c|}{IFRNet \cite{kong2022ifrnet}} & \multicolumn{3}{c|}{AMT-S \cite{li2023amt}} & \multicolumn{3}{c|}{EMA-VFI \cite{zhang2023extracting}} \\
    \cmidrule(r){2-4}
    \cmidrule(r){5-7} 
    \cmidrule(r){8-10}
    \cmidrule(r){11-13}
    & $[T]$ & $[D]$ & $[D]_{u}$ & $[T]$ & $[D]$ & $[D]_{u}$ & $[T]$ & $[D]$ & $[D]_{u}$ & $[T]$ & $[D]$ & $[D]_{u}$\\
    \cmidrule(r){2-4}
    \cmidrule(r){5-7} 
    \cmidrule(r){8-10}
    \cmidrule(r){11-13}    
    PSNR$\uparrow$ & 35.61 & \bestscore{36.04} & \fakescore{35.18} & 35.80 & \bestscore{36.26} & \fakescore{35.14} & 35.97 & \bestscore{36.56} & \fakescore{35.21} & 36.50 & \bestscore{37.13} & \fakescore{36.21} \\
    SSIM$\uparrow$ & 0.978 & \bestscore{0.979} & \fakescore{0.976} & 0.979 & \bestscore{0.981} & \fakescore{0.977} & 0.980 & \bestscore{0.982} & \fakescore{0.977} & 0.982 & \bestscore{0.983} & \fakescore{0.981} \\
    LPIPS$\downarrow$ & 0.022 & \bestscore{0.022} & 0.023 & 0.020 & \bestscore{0.019} & 0.021 & 0.021 & \bestscore{0.020} & 0.023 & 0.020 & \bestscore{0.019} & 0.020 \\
    NIQE$\downarrow$ & 5.249 & 5.225 & \bestscore{5.224} & 5.256 & 5.245 & \bestscore{5.225} & 5.308 & 5.293 & \bestscore{5.288} & 5.372 & 5.343 & \bestscore{5.335} \\
    \bottomrule
  \end{tabular}
\end{center}
\end{table*}

\section{Additional Experiments and Analysis}
\label{sec:additional_experiments}

\subsection*{Comparison of the fixed-time setting} 
While the benefits of our proposed disambiguation strategies are best demonstrated on arbitrary-time VFI models, they actually improve the performance of fixed-time models as well. Using RIFE \cite{huang2022real} as a representative example, we extend our comparison to the fixed-time training paradigm, depicted in \cref{fig:additional_qualitative_comparison}. The label $[T]$ RIFE (Tri) refers to the model trained on the triplet dataset from Vimeo90K \cite{xue2019video} employing time indexing. Conversely, $[D]$ RIFE (Tri) indicates training on the same triplet dataset but utilizing our distance indexing approach. Both $[D]$ RIFE and $[D,R]$ RIFE models are trained on the septuplet dataset, consistent with our earlier comparison. Despite being trained on varied datasets, it is evident that the arbitrary time model outperforms the fixed time model. However, the efficacy of distance indexing appears restrained within the fixed-time training paradigm. This limitation stems from the fact that deriving distance representation solely from the middle frame yields a sparse distribution, making it challenging for the network to grasp the nuances of distance. We delve deeper into the quantitative analysis of these findings in \cref{tab:vimeo90k_triplet}. Compared to training arbitrary time models on the septuplet dataset, the advantages of distance indexing become notably decreased when training fixed time models on the triplet dataset.

\subsection*{Two channel scheme} Comparison results with the two channel scheme, \ie, $D_t$ with $\bm{x}$ and $\bm{y}$ directions, are shown in \cref{tab:vimeo90k_septuplet_appendix}. The observations are as follows: (1) $[D_{x,y}]_u$ outperforms $[D]_u$, which makes sense since it accounts for both speed and direction. (2) $[D,R]_u$ performs better than $[D_{x,y}]_u$. Our speculation is that $[R]$ benefits not only from addressing the directional ambiguity but also from reducing the prediction difficulty via divide-and-conquer. (3) $[D_{x,y},R]_u$ does not exceed $[D,R]_u$, showing that the iterative formulation is sufficient to resolve the directional ambiguity. $[D_{x,y}]_u$ has the potential to learn different trajectories but cannot realize that potential since the trajectory distribution within a short time is not sufficiently diverse. However, $[D_{x,y}]_u$ only involves increasing the input channels and does not need to run iteratively. Thus, $[D_{x,y}]_u$ is a better choice for fast/lightweight frame interpolation.

\begin{table}[!t]
\caption{Comparison with $[D_{x,y}]$ scheme on Vimeo90K septuplet dataset. $[D_{x,y}]$ denotes training with two channel distance map with $\bm{x}$ and $\bm{y}$ directions. $[\cdot]_u$ denotes inference with uniform maps. 
}
\label{tab:vimeo90k_septuplet_appendix}
\setlength{\tabcolsep}{6pt}
\begin{center}
  \begin{tabular}{lccccc}
    \toprule
    & $[T]$ & $[D]_u$ & $[D,R]_u$ & $[D_{x,y}]_u$ & $[D_{x,y},R]_u$ \\
    \midrule
    LPIPS$\downarrow$ & 0.105 & 0.092 & \bestscore{0.086} & 0.091 & 0.087 \\
    NIQE$\downarrow$ & 6.663 & 6.344 & \bestscore{6.220} & 6.296 & \bestscore{6.220} \\
    \bottomrule
  \end{tabular}
\end{center}
\end{table}

\subsection*{Why did $[R]$ fail on AMT-S?} As compared to the speed ambiguity, the directional ambiguity only has a minor impact to the interpolation quality due to the short time span between frames. Merely employing iterative reference-based estimation without tackling speed ambiguity $[T,R]$ can result in the accumulation of inaccuracies. This phenomenon is especially evident in AMT-S, which we attribute to its scaled lookup operation of bidirectional 4D correlation volumes. Cumulative errors are exacerbated in inaccurate iterative lookup operations.

\subsection*{Why NIQE performs better with uniform maps?} Uniform maps tend to yield better results due to smoothing. Nonuniform maps consist of unavoidable inaccuracies from flow estimation, which may introduce unnatural details, resulting in worse NIQE scores but are not noticeable to human perception.

\begin{figure*}[!t]
    \begin{center}
    \includegraphics[width=\linewidth]{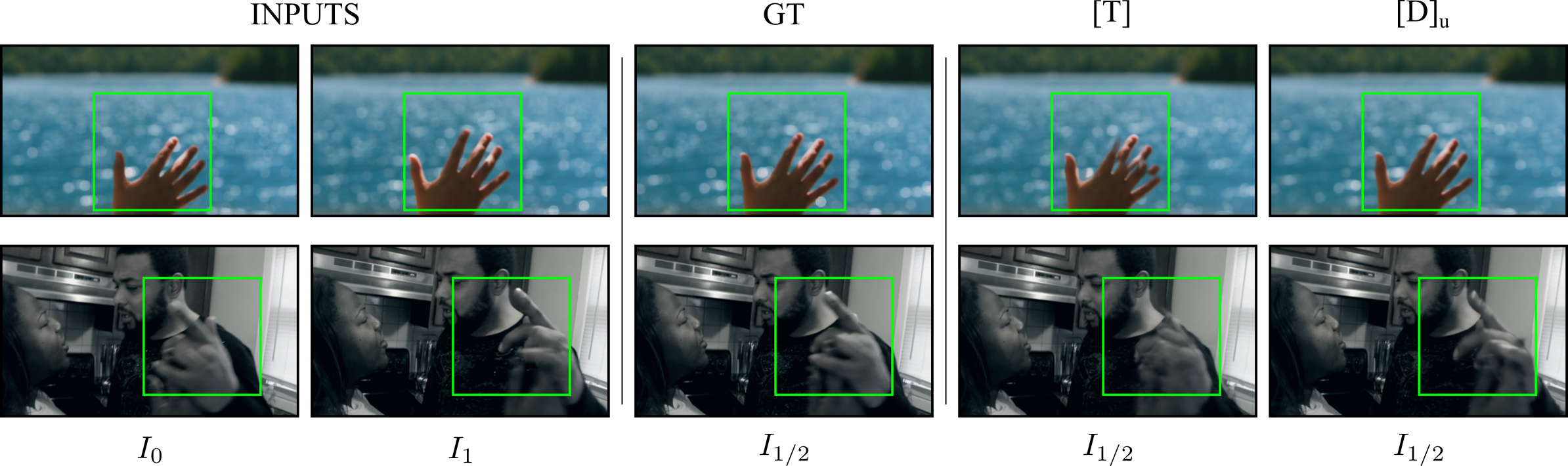}
    \end{center}
    \caption{\extension{Comparison on Vimeo90K Septuplet with GiMM-VFI~\cite{guo2024generalizable}.}}
    \label{fig:gimmvfi}
\end{figure*}

\begin{table}[!t]
\caption{\extension{Comparison on Vimeo90K Septuplet with GiMM-VFI~\cite{guo2024generalizable} and BiM-VFI~\cite{seo2025bim}.}}
\label{tab:vimeo90k_gimmvfi}
\setlength{\tabcolsep}{2pt}
\centering
\begin{tabular}{l|cccc}
\toprule
 & BiM-VFI & GiMM-VFI $[T]$ & GiMM-VFI $[D]$ & GiMM-VFI $[D]_u$ \\
\midrule
PSNR$\uparrow$ & \fakescore{26.83} & 28.36 & \bestscore{29.36} & \fakescore{27.58} \\
SSIM$\uparrow$ & \fakescore{0.893} & 0.930 & \bestscore{0.943} & \fakescore{0.919} \\
LPIPS$\downarrow$ & 0.070 & 0.046 & \bestscore{0.041} & 0.045 \\
NIQE$\downarrow$ & 6.009 & 6.107 & 6.004 & \bestscore{5.998} \\
\bottomrule
\end{tabular}
\end{table}

\subsection*{\extension{Evaluation on strong mainstream baselines}} 
\extension{We further consider recent strong models based on mainstream VFI architectures, including GIMM-VFI~\cite{guo2024generalizable} and BiM-VFI~\cite{seo2025bim}. Since BiM-VFI can be viewed as a direct extension of our line of work, which introduces a new motion description (Bidirectional Motion Fields) built upon distance maps, we focus our evaluation on GIMM-VFI for a representative comparison.
As shown in~\cref{tab:vimeo90k_gimmvfi}, training GIMM-VFI with distance indexing consistently improves performance over its original time-indexed version. We also provide a qualitative comparison in~\cref{fig:gimmvfi}. These results demonstrate that the proposed strategy remains effective even when applied to strong, modern architectures.}

\begin{table}[!t]
    \centering
    \caption{\extension{Comparison on Adobe240~\cite{su2017deep} for $\times$8 interpolation with RIFE~\cite{huang2022real}.}}
    \label{tab:adobe240}
    \setlength{\tabcolsep}{12pt}
    \begin{tabular}{lccc}
        \toprule
        & $[T]$ & $[D]_{u}$ & $[D,R]_{u}$\\
        \midrule
        PSNR $\uparrow$ & 30.24 & \bestscore{\fakescore{30.47}} & \fakescore{30.30} \\
        SSIM $\uparrow$ & \bestscore{0.939} & \fakescore{0.938} & \fakescore{0.937} \\
        LPIPS $\downarrow$ & 0.073 & 0.057 & \bestscore{0.054} \\
        NIQE $\downarrow$ & 5.206 & 4.974 & \bestscore{4.907} \\
        \bottomrule
    \end{tabular}
\end{table}

\begin{table}[!t]
    \centering
    \caption{\extension{Comparison on X4K1000FPS~\cite{sim2021xvfi} for $\times$8 interpolation with RIFE~\cite{huang2022real}.}}
    \label{tab:x4k1000fps}
    \setlength{\tabcolsep}{12pt}
    \begin{tabular}{lccc}
        \toprule
        & $[T]$ & $[D]_{u}$ & $[D,R]_{u}$ \\
        \midrule
        PSNR $\uparrow$ & 36.36 & \bestscore{\fakescore{36.80}} & \fakescore{36.26} \\
        SSIM $\uparrow$ & \bestscore{0.967} & \fakescore{0.964} & \fakescore{0.964} \\
        LPIPS $\downarrow$ & 0.040 & 0.032 & \bestscore{0.032} \\
        NIQE $\downarrow$ & 7.130 & 6.936 & \bestscore{6.924} \\
        \bottomrule
    \end{tabular}
\end{table}

\subsection*{\extension{More results on other benchmarks}}
\label{sec:more_results}
\extension{\cref{tab:adobe240} and \cref{tab:x4k1000fps} show the results of RIFE~\cite{huang2022real} on Adobe240~\cite{su2017deep} and X4K1000FPS~\cite{sim2021xvfi} for $\times$8 interpolation respectively, using uniform maps. Distance indexing $[D]_{u}$ and iterative reference-based estimation $[R]_{u}$ strategies can consistently help improve the perceptual quality. In addition, it is noteworthy that $[D]_{u}$ is better than $[T]$ in terms of the pixel-centric metrics like PSNR, showing that the constant speed assumption (uniform distance maps) holds well on these two easier benchmarks.}

\begin{table*}[!t]
\centering
\caption{\extensionvtwo{Performance comparison across motion-pattern categories for representative VFI models.
The evaluation set is partitioned into four motion types based on optical-flow-derived velocity statistics: \textit{acceleration}, \textit{deceleration}, \textit{approximately constant-speed}, and \textit{complex mixed acceleration/deceleration}.
Within each model and motion type, we compare uniform time indexing $[T]$ and ground-truth distance indexing $[D]$.
The better result between $[T]$ and $[D]$ is highlighted in \textbf{bold}.}}
\label{tab:system_analysis}
\setlength{\tabcolsep}{12pt}
\renewcommand{\arraystretch}{1.15}
\begin{tabular}{lllcccc}
\toprule
Model & Motion Type & Cat. & PSNR $\uparrow$ & SSIM $\uparrow$ & LPIPS $\downarrow$ & NIQE $\downarrow$ \\
\midrule

\multirow{8}{*}{RIFE~\cite{huang2022real}}
& \multirow{2}{*}{Acceleration}
& $[T]$ & 28.2145 & 0.9117 & 0.1015 & 6.7908 \\
& & $[D]$ & \textbf{29.2395} & \textbf{0.9290} & \textbf{0.0899} & \textbf{6.5830} \\
\cmidrule(lr){2-7}
& \multirow{2}{*}{Complex}
& $[T]$ & 28.4378 & 0.9126 & 0.1041 & 6.5606 \\
& & $[D]$ & \textbf{29.3960} & \textbf{0.9292} & \textbf{0.0912} & \textbf{6.3613} \\
\cmidrule(lr){2-7}
& \multirow{2}{*}{Constant}
& $[T]$ & 29.4517 & 0.9235 & 0.1017 & 6.1813 \\
& & $[D]$ & \textbf{30.2031} & \textbf{0.9347} & \textbf{0.0848} & \textbf{5.9581} \\
\cmidrule(lr){2-7}
& \multirow{2}{*}{Deceleration}
& $[T]$ & 27.8605 & 0.9088 & 0.1042 & 6.6731 \\
& & $[D]$ & \textbf{28.8007} & \textbf{0.9262} & \textbf{0.0971} & \textbf{6.5364} \\
\midrule

\multirow{8}{*}{IFRNet~\cite{kong2022ifrnet}}
& \multirow{2}{*}{Acceleration}
& $[T]$ & 28.2330 & 0.9146 & 0.0869 & 6.5339 \\
&  & $[D]$ & \textbf{29.2812} & \textbf{0.9314} & \textbf{0.0785} & \textbf{6.4588} \\
\cmidrule(lr){2-7}
& \multirow{2}{*}{Complex}
& $[T]$ & 28.4757 & 0.9161 & 0.0883 & 6.3191 \\
&  & $[D]$ & \textbf{29.4851} & \textbf{0.9322} & \textbf{0.0782} & \textbf{6.2284} \\
\cmidrule(lr){2-7}
& \multirow{2}{*}{Constant}
& $[T]$ & 29.4212 & 0.9259 & 0.0850 & 6.0017 \\
&  & $[D]$ & \textbf{30.2535} & \textbf{0.9370} & \textbf{0.0716} & \textbf{5.8317} \\
\cmidrule(lr){2-7}
& \multirow{2}{*}{Deceleration}
& $[T]$ & 27.8972 & 0.9123 & 0.0900 & 6.4507 \\
&  & $[D]$ & \textbf{28.8555} & \textbf{0.9296} & \textbf{0.0841} & \textbf{6.3932} \\
\midrule

\multirow{8}{*}{AMT-S~\cite{li2023amt}}
& \multirow{2}{*}{Acceleration}
& $[T]$ & 28.4748 & 0.9190 & 0.1008 & 6.9927 \\
&  & $[D]$ & \textbf{29.6034} & \textbf{0.9368} & \textbf{0.0855} & \textbf{6.7747} \\
\cmidrule(lr){2-7}
& \multirow{2}{*}{Complex}
& $[T]$ & 28.7413 & 0.9211 & 0.1013 & 6.7628 \\
&  & $[D]$ & \textbf{29.8756} & \textbf{0.9386} & \textbf{0.0832} & \textbf{6.5250} \\
\cmidrule(lr){2-7}
& \multirow{2}{*}{Constant}
& $[T]$ & 29.9110 & 0.9346 & 0.0934 & 6.4211 \\
&  & $[D]$ & \textbf{30.7599} & \textbf{0.9457} & \textbf{0.0734} & \textbf{6.0756} \\
\cmidrule(lr){2-7}
& \multirow{2}{*}{Deceleration}
& $[T]$ & 28.1143 & 0.9160 & 0.1033 & 6.8830 \\
&  & $[D]$ & \textbf{29.1998} & \textbf{0.9349} & \textbf{0.0917} & \textbf{6.7289} \\
\midrule

\multirow{8}{*}{EMA-VFI~\cite{zhang2023extracting}}
& \multirow{2}{*}{Acceleration}
& $[T]$ & 29.3060 & 0.9263 & 0.0871 & 6.8722 \\
&  & $[D]$ & \textbf{30.2554} & \textbf{0.9410} & \textbf{0.0781} & \textbf{6.6792} \\
\cmidrule(lr){2-7}
& \multirow{2}{*}{Complex}
& $[T]$ & 29.7016 & 0.9302 & 0.0839 & 6.6009 \\
&  & $[D]$ & \textbf{30.5960} & \textbf{0.9437} & \textbf{0.0746} & \textbf{6.4067} \\
\cmidrule(lr){2-7}
& \multirow{2}{*}{Constant}
& $[T]$ & 31.0516 & 0.9437 & 0.0708 & 6.1619 \\
&  & $[D]$ & \textbf{31.6974} & \textbf{0.9511} & \textbf{0.0619} & \textbf{5.9517} \\
\cmidrule(lr){2-7}
& \multirow{2}{*}{Deceleration}
& $[T]$ & 28.9662 & 0.9245 & 0.0909 & 6.7890 \\
&  & $[D]$ & \textbf{29.8220} & \textbf{0.9395} & \textbf{0.0838} & \textbf{6.6091} \\

\bottomrule
\end{tabular}
\end{table*}

\begin{figure}[!t]
    \begin{center}
    \includegraphics[width=\linewidth]{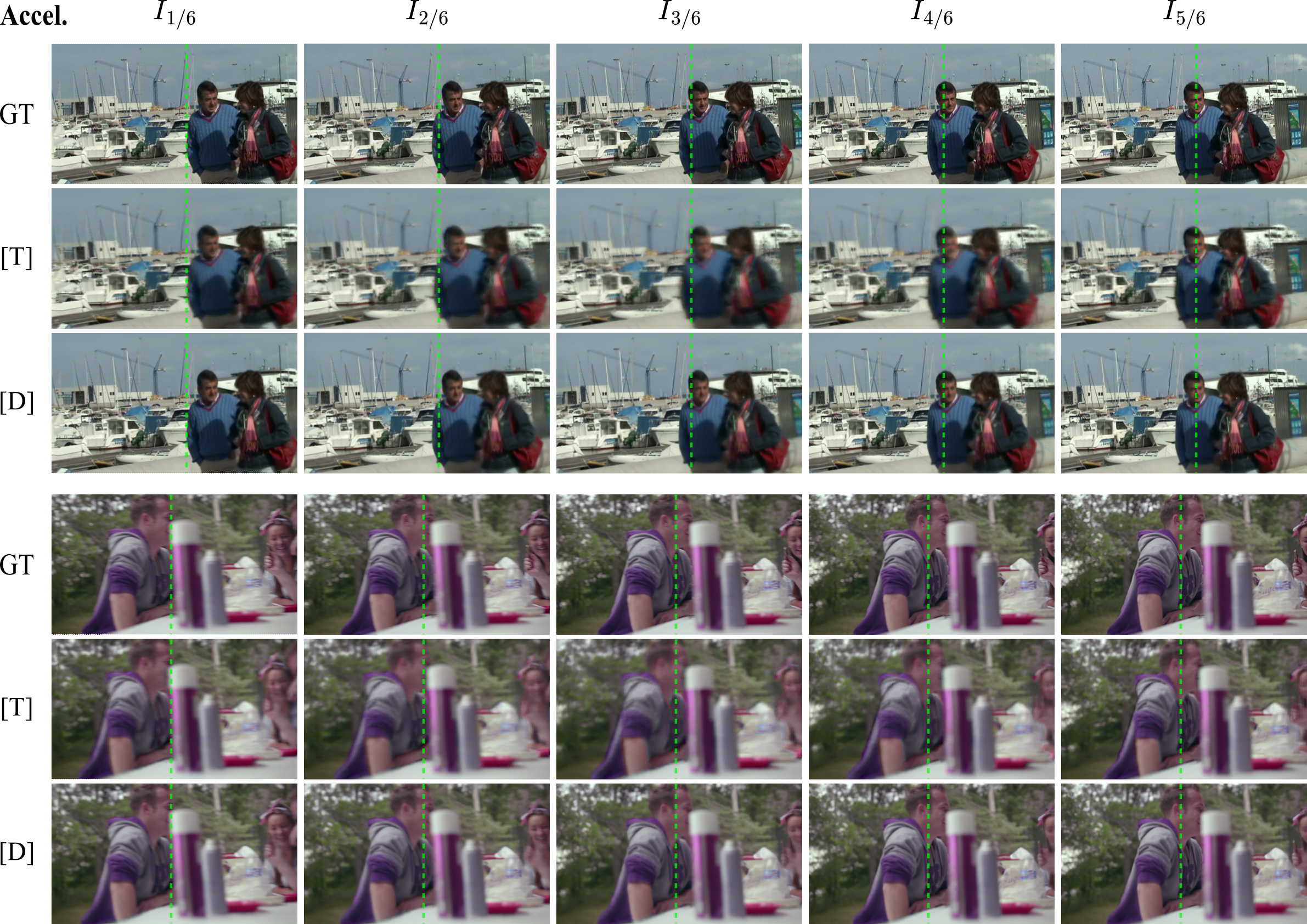}
    \end{center}
    \caption{\extensionvtwo{Qualitative comparison on \textit{acceleration} motion patterns. Distance indexing ($[D]$) better preserves motion details and reduces blur artifacts compared to time indexing ($[T]$), consistent with the systematic results in~\cref{tab:system_analysis}.}}
    \label{fig:system_analysis_accel}
\end{figure}

\begin{figure}[!t]
    \begin{center}
    \includegraphics[width=\linewidth]{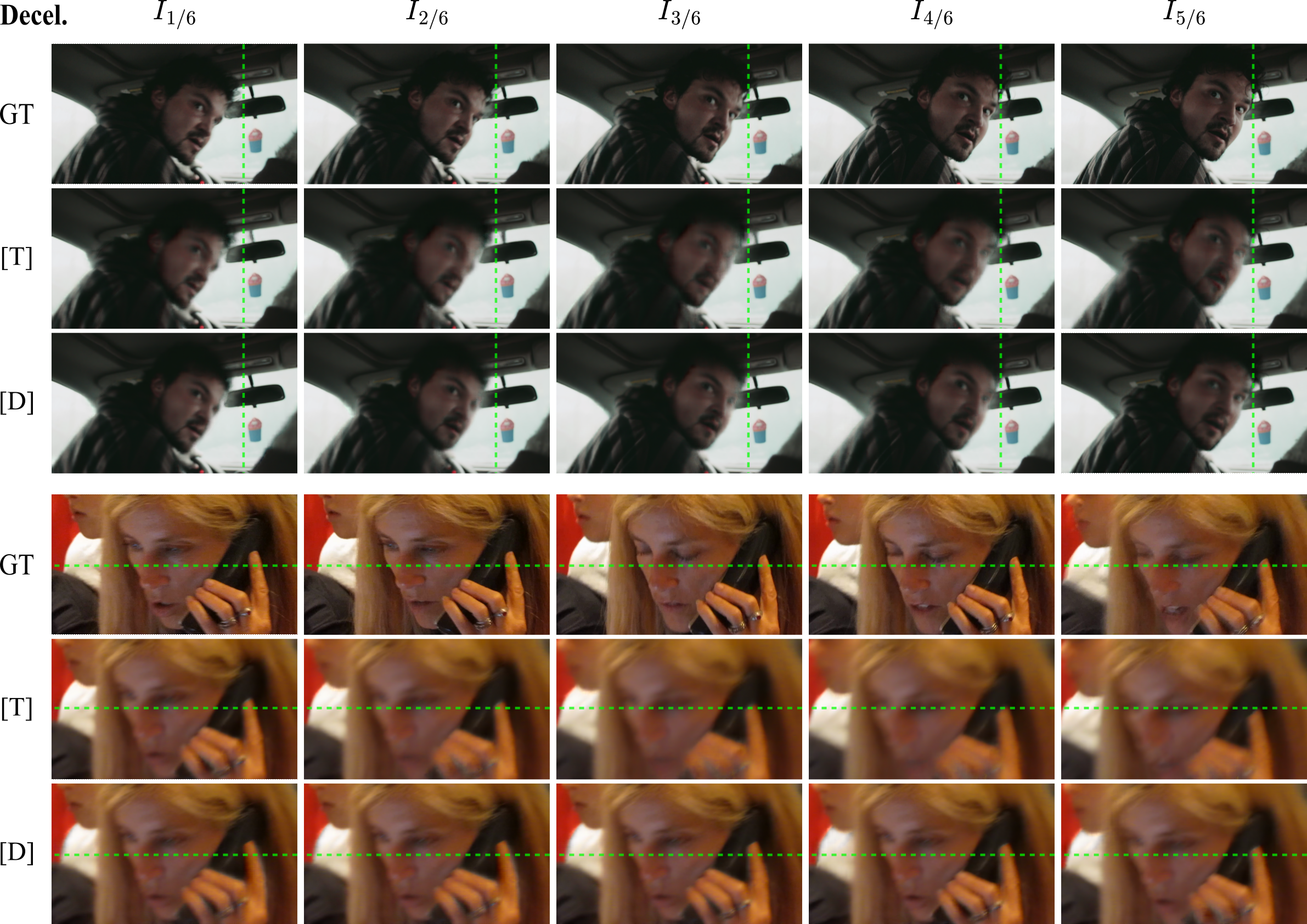}
    \end{center}
    \caption{\extensionvtwo{Qualitative comparison on \textit{deceleration} motion patterns. Distance indexing ($[D]$) better preserves motion details and reduces blur artifacts compared to time indexing ($[T]$), consistent with the systematic results in~\cref{tab:system_analysis}.}}
    \label{fig:system_analysis_decel}
\end{figure}

\subsection*{\extensionvtwo{Systematic analysis on motion patterns}}

\extensionvtwo{We provide a more direct and systematic analysis to quantify how distance indexing differentiates motion patterns here.
Specifically, we categorize samples in the evaluation dataset of Vimeo90K Septuplet into four motion-pattern groups based on the temporal variation of optical flow magnitude (\ie, velocity statistics): \textit{acceleration}, \textit{deceleration}, \textit{approximately constant-speed}, and \textit{complex mixed acceleration/deceleration}. 
For each sample, we compute the optical flow magnitude for the intermediate frames and analyze its temporal trend. We then assign it to one of the above motion categories according to whether the velocity profile is increasing, decreasing, nearly constant, or exhibits mixed non-monotonic changes. We evaluate representative VFI models under uniform time indexing ($[T]$) and ``ground-truth'' distance indexing ($[D]$) within each motion category.
The goal of this analysis is to verify whether distance indexing can consistently improve interpolation quality under diverse velocity profiles.}

\extensionvtwo{As shown in~\cref{tab:system_analysis}, using accurate distance indexing consistently improves performance across all motion-pattern groups and across all evaluated models. 
Notably, the gains remain stable not only for approximately constant-speed motion, but also for accelerating and decelerating cases where velocity changes over time.
This suggests that distance indexing provides a more expressive and physically meaningful motion parameterization, enabling the interpolated results to better align with diverse motion patterns.}

\extensionvtwo{In addition to the quantitative comparisons, we further provide representative qualitative examples for the acceleration and deceleration cases in~\cref{fig:system_analysis_accel,fig:system_analysis_decel}, where the improvements in motion behavior and interpolation quality are visually consistent with the quantitative findings.}

\section{\extension{Implementation Details}}
\label{sec:implementation}
\extension{For VFI models that explicitly take the temporal index $t$ as input, we directly replace $t$ with the corresponding distance index $D_t$, without modifying the network architecture. For models that implicitly target a fixed interpolation time (\eg, predicting $I_{0.5}$), we concatenate the distance map $D_t$ with the RGB input frames as an additional channel. In practice, this only requires expanding the input channels of the first encoder layer, while keeping all network modules unchanged. For models with a shared encoder, the distance map is concatenated consistently with both input frames.}

\extension{The same strategy applies to the iterative reference-based estimation setting ($[R]$), where the reference RGB image and distance map are concatenated to replace the original time indexing input. All changes are confined to the input representation and do not involve architectural redesign, but only a minimal expansion of the input channels.}

\begin{figure*}[!t]
    \begin{center}
        \includegraphics[width=.9\linewidth]{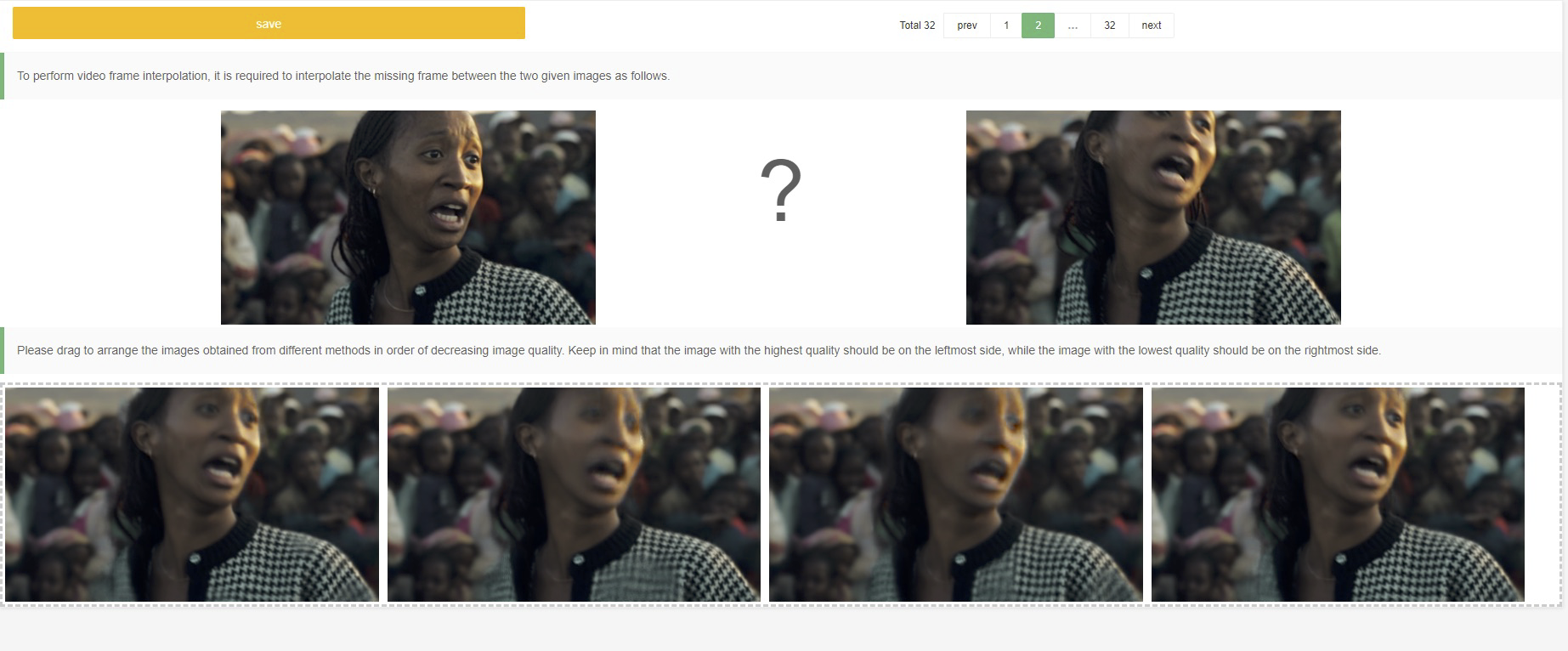}
    \end{center}
\caption{User interface of user study.}
\label{fig:ui}
\end{figure*}

\section{User Study UI}
\label{sec:ui}
As shown in \cref{fig:ui}, we initially presented users with the input starting and ending frames. Subsequently, the results from each model's four distinct variants were displayed anonymously in a sequence, with the order shuffled for each presentation. Users were tasked with reordering the images by dragging them, placing them from left to right based on their perceived quality, \ie, the best image on the extreme left and the least preferred on the far right.

\section{Demo}
\label{sec:demo}
We have included a video demo (available in supplementary materials named ``supp.mp4'') to intuitively showcase the enhanced quality achieved through our strategies. The video further illustrates the idea of manipulating object interpolations and provides a guide on using the related web application.
}

\bibliographystyle{IEEEtran}
\bibliography{main}



\begin{IEEEbiography}[{\includegraphics[width=1in,height=1.25in,clip,keepaspectratio]{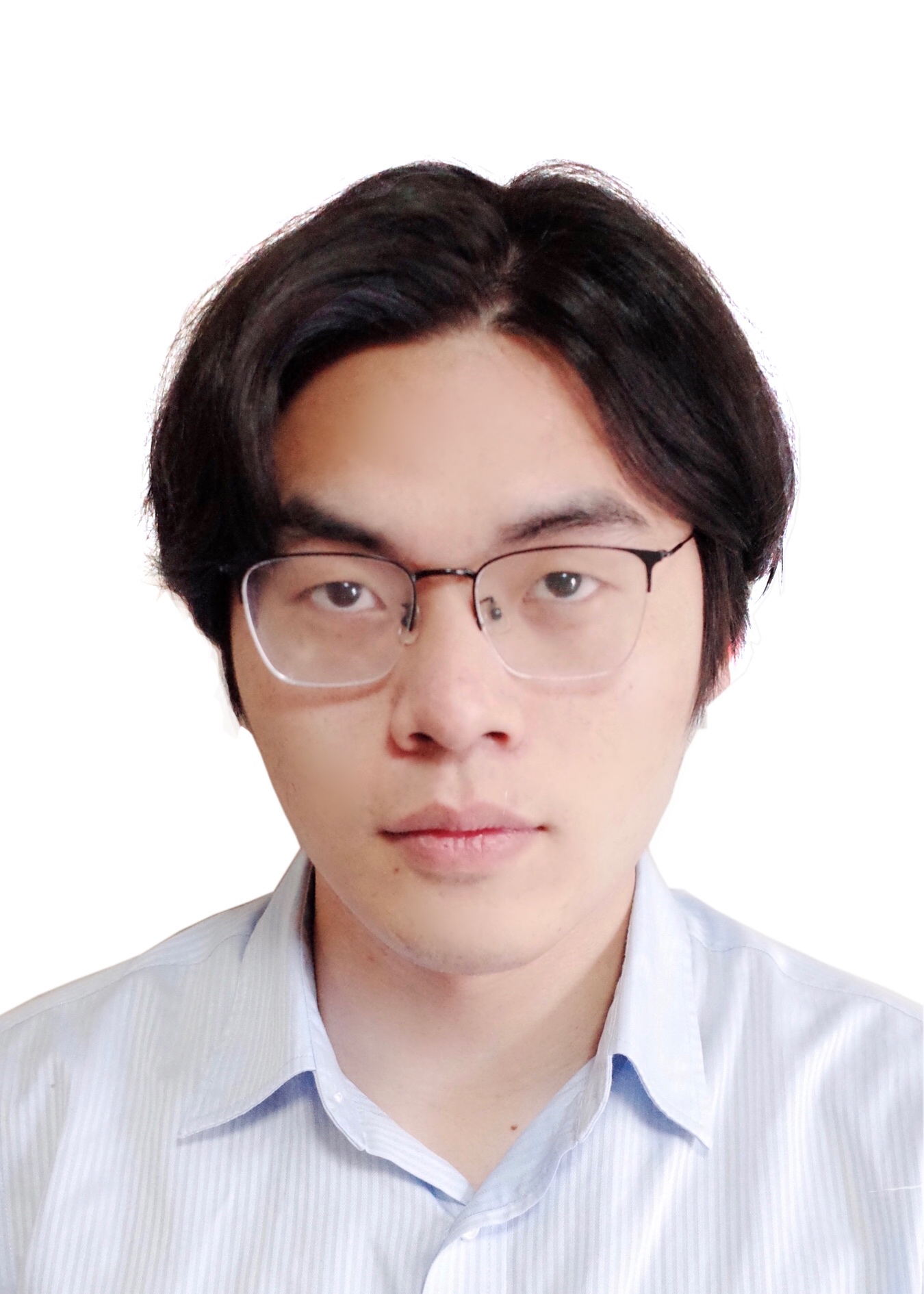}}]{Zhihang Zhong}
is an Associate Professor with the School of Artificial Intelligence, Shanghai Jiao Tong University, since 2026. He has also been a Researcher at Shanghai AI Laboratory since 2023. In 2018, he received the B.E. degree in mechatronics engineering with Chu Kochen Honors from Zhejiang University. He received the M.E. degree in precision engineering and the Ph.D. degree in computer science from the University of Tokyo in 2020 and 2023, respectively. He has published papers in top-tier conferences and journals, such as CVPR, ECCV, ICCV, IJCV, \etc. His current research interests include neural rendering, spatial agents, and computational photography.
\end{IEEEbiography}

\begin{IEEEbiography}
[{\includegraphics[width=1in,height=1.25in,clip,keepaspectratio]{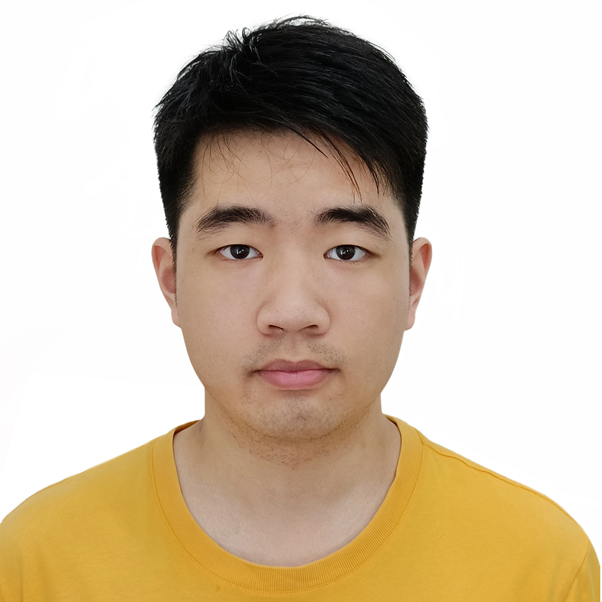}}]{Yiming Zhang} is currently a master student at Cornell University and works as an intern at Shanghai AI Laboratory. He received his B.E. degree in Electrical and Computer Engineering from Shanghai Jiao Tong University in 2023. His research interests include diffusion generative models, multi-agent intelligence, and trajectory analysis. 
\end{IEEEbiography}

\begin{IEEEbiography}
[{\includegraphics[width=1in,height=1.25in,clip,keepaspectratio]{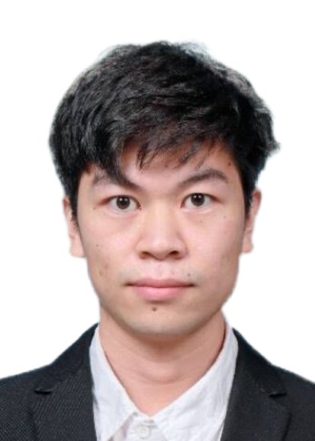}}]{Wei Wang}is currently a Research Engineer at Shanghai Artificial Intelligence Laboratory. His research interests include object detection, human pose estimation and 3D reconstruction. He received the B.S.  from University of Chinese Academy of Sciences in 2020, and the M.S. degree from the Institute of Information Engineering, Chinese Academy of Sciences, Beijing, China, in 2023.
\end{IEEEbiography}

\begin{IEEEbiography}
[{\includegraphics[width=1in,height=1.25in,clip,keepaspectratio]{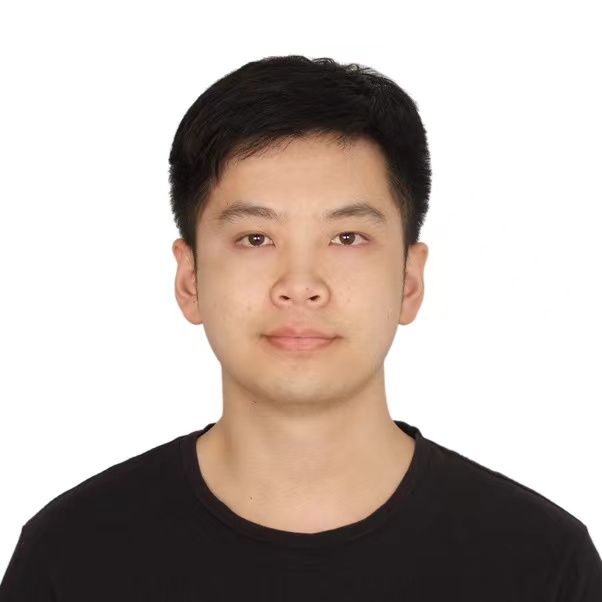}}]{Xiao Sun}is a scientist at Shanghai AI Laboratory, where he leads diverse R\&D groups on AI for sports, healthcare, and robotics. Before that, he served as a Senior Researcher at the Visual Computing Group, Microsoft Research Asia (MSRA), from Feb. 2016 to Jul. 2022. Xiao received the B.S. and M.S. degrees in Information Engineering from South China University of Technology, China, in 2011 and 2014, respectively. His research interests include computer vision, machine learning, and computer graphics.
\end{IEEEbiography}

\begin{IEEEbiography}
[{\includegraphics[width=1in,height=1.25in,clip,keepaspectratio]{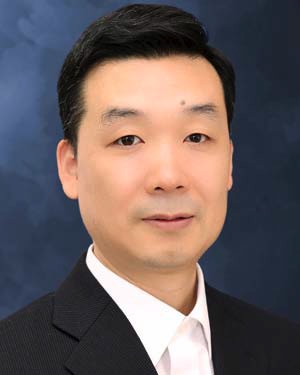}}]{Yu Qiao} is the lead scientist with Shanghai Artificial Intelligence Laboratory, a researcher, and the honored director of the Multimedia Laboratory with Shenzhen Institutes of Advanced Technology, Chinese Academy of Sciences (SIAT). He served as an assistant professor with the Graduate School of Information Science and Technology, University of Tokyo from 2009 to 2010. He has been working on deep learning since 2006, and he is one of the earliest people to introduce deep learning to video understanding. He and his team invented center loss and temporal segment networks. He has published more than 400 articles in top-tier conferences and journals in computer science with more than 80,000 citations. He received the CVPR 2023 Best Paper Award and AAAI 2021 Outstanding Paper Award. His research interests revolve around foundation models, computer vision, deep learning, robotics, and AI applications.
\end{IEEEbiography}

\begin{IEEEbiography}
[{\includegraphics[width=1in,height=1.25in,clip,keepaspectratio]{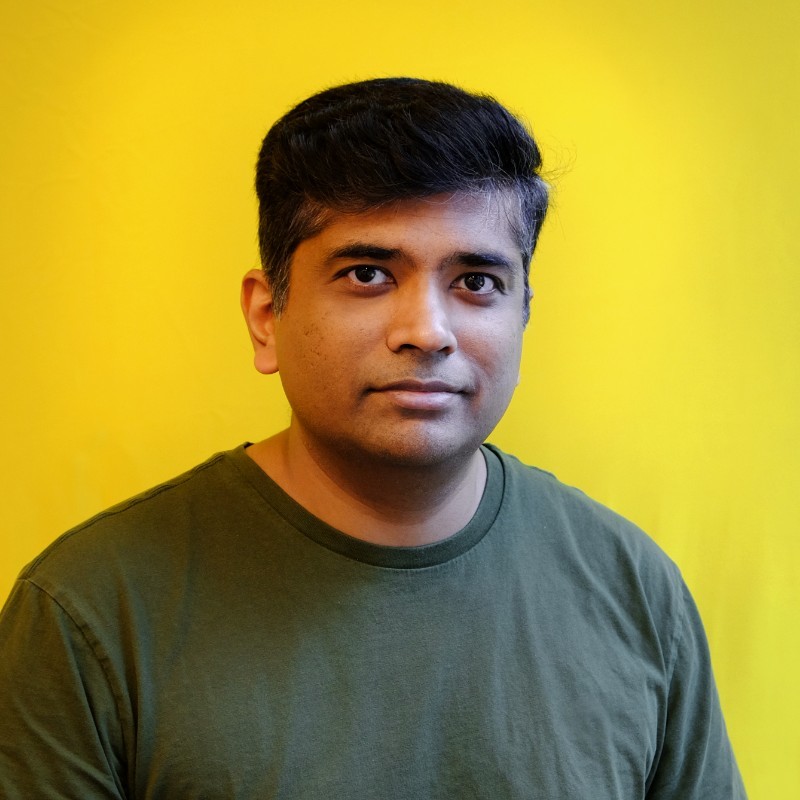}}]{Gurunandan Krishnan} is an SVP of OtoNexus Medical Technologies. He obtained his master's degree from Columbia University and bachelor's degree from Visvesvaraya Technological University. His research interests are in Computer Vision, Computer Graphics, and Computational Imaging. He has published papers in top-tier conferences and journals, such as Siggraph, CVPR, \etc.
\end{IEEEbiography}

\begin{IEEEbiography}
[{\includegraphics[width=1in,height=1.25in,clip,keepaspectratio]{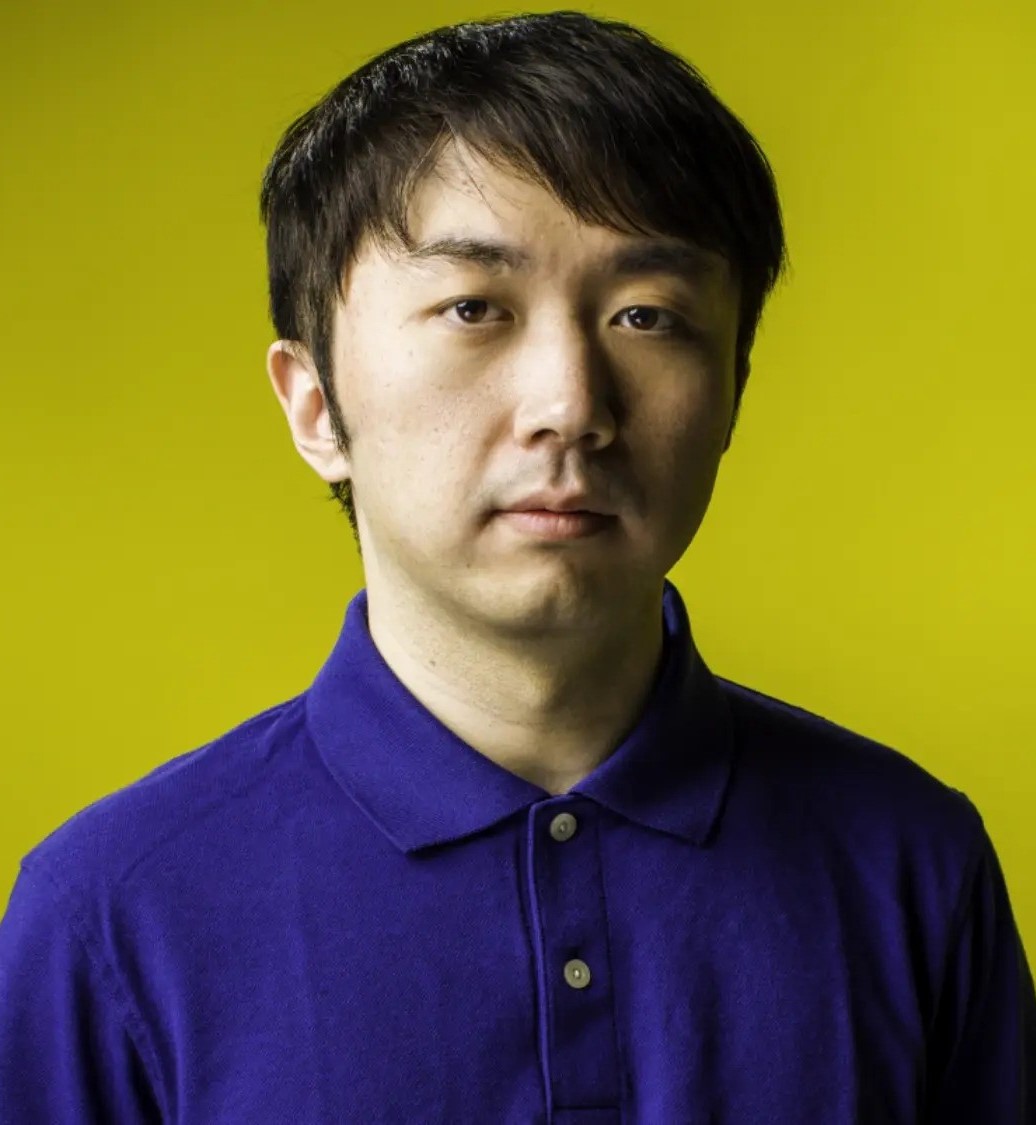}}]{Sizhuo Ma}is a Senior Research Scientist at Snap Inc. He obtained a Ph.D. degree from University of Wisconsin-Madison in 2022. He received his bachelor's degree from Shanghai Jiao Tong University. His research interests include computational imaging, computational photography and low-level vision. He has published papers at prestigious journals and conferences including CVPR, ECCV, Siggraph, MobiCom, ISMAR, \etc. 
\end{IEEEbiography}

\begin{IEEEbiography}
[{\includegraphics[width=1in,height=1.25in,clip,keepaspectratio]{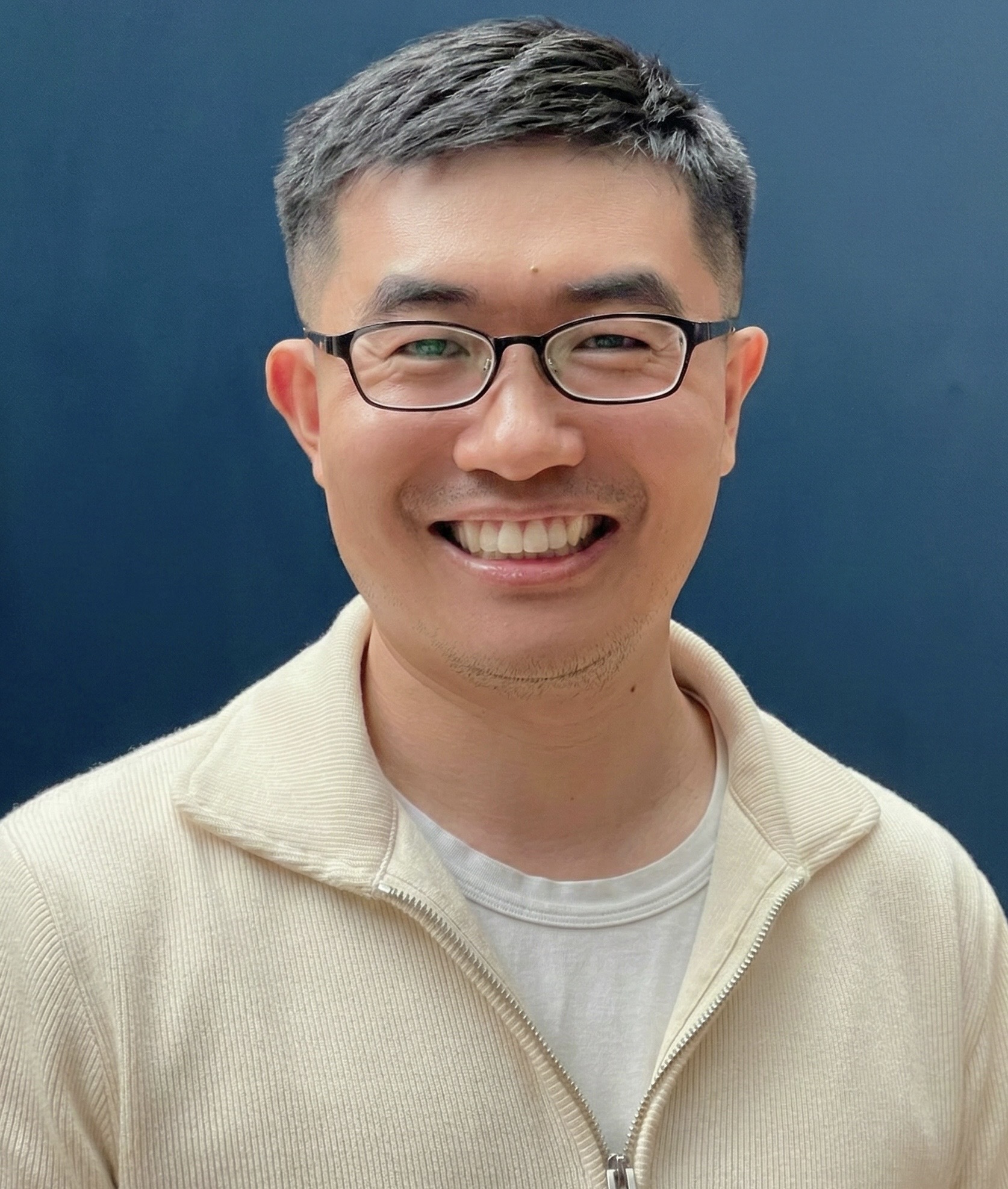}}]{Jian Wang} is a Staff Research Scientist at Snap Inc., focusing on computational photography and imaging. He has published in top-tier venues such as CVPR, MobiCom, and SIGGRAPH, and has contributed numerous features to production. He has received Best Paper awards at SIGGRAPH Asia 2024 and the 4th IEEE International Workshop on Computational Cameras and Displays, as well as Best Poster awards at the IEEE Conference on Computational Photography 2022 and the Responsible Imaging workshop at ICCV 2025. He has served as an Area Chair for CVPR, NeurIPS, ICLR, ICML, \etc. Jian holds a Ph.D. from Carnegie Mellon University.
\end{IEEEbiography}



\vfill

\end{document}